\documentclass[review]{elsarticle}

\usepackage{lineno,hyperref}
\usepackage{multicol}
\usepackage{multirow}
\usepackage{bigstrut}
\usepackage{amsmath}
\usepackage{adjustbox}
\usepackage[table,xcdraw]{xcolor}

\begin{document}

\begin{frontmatter}

\title{Visual Explanation of Black-Box Model :~Similarity Difference and Uniqueness (SIDU) Method}  

\author{\hspace{-0.55cm}Satya M. Muddamsetty$^{1}$~ Mohammad N. S. Jahromi$^{1}$~ Andreea E. Ciontos$^{2}$\\~ Laura M. Fenoy$^{3}$~ Thomas B. Moeslund$^{1}$}
\address{$^{1}$Visual Analysis and Perception Laboratory (VAP), Aalborg University, Aalborg, Denmark\\
$^{2}$Department of Material and Production, Aalborg University, Aalborg, Denmark\\
$^{3}$ Yodaway, Aalborg, Denmark}





\begin{abstract}
Explainable Artificial Intelligence (XAI) has in recent years become a well-suited framework to generate human understandable explanations of 'black- box' models. In this paper, a novel XAI visual explanation algorithm known as the \textbf{Similarity Difference and Uniqueness} (SIDU) method that can effectively localize entire object regions responsible for prediction is presented in full detail. The SIDU algorithm robustness and effectiveness is analyzed through various computational and human subject experiments. In particular, the SIDU algorithm is assessed using three different types of evaluations (Application, Human and Functionally-Grounded) to demonstrate its superior performance. The robustness of SIDU is further studied in the presence of adversarial attack on 'black-box' models to better understand its performance. Our code is available at:~~\url{https://github.com/satyamahesh84/SIDU_XAI_CODE}.
\end{abstract}


\end{frontmatter}


\section{Introduction}
\begin{figure*} [t!]
\begin{centering}
\includegraphics[width= 0.23\textwidth, height = 2.5cm]{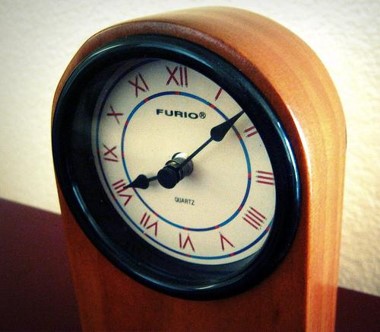}
\hspace{0.01\columnwidth}
\includegraphics[width= 0.23\textwidth, height = 2.5cm]{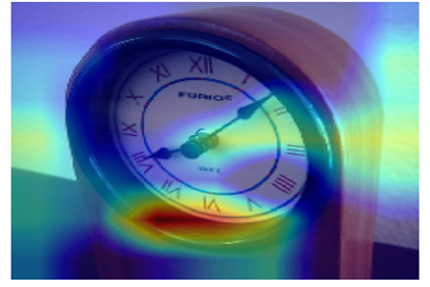}
\hspace{0.01\columnwidth}
\includegraphics[width= 0.23\textwidth, height = 2.5cm]{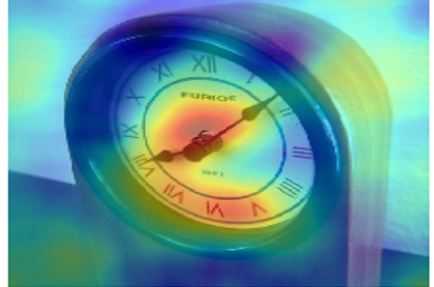}
\hspace{0.01\columnwidth}
\includegraphics[width= 0.23\textwidth, height = 2.5cm]{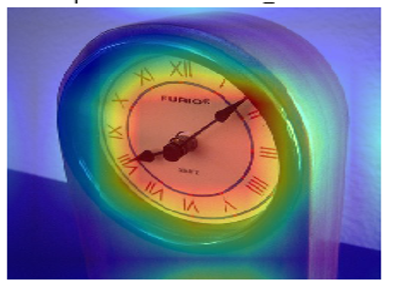}

\par\end{centering}
\begin{centering}
\vspace{1mm}
\par\end{centering}
\centering{}
(a) Original image\hspace{0.3cm} b) GRAD-CAM\hspace{0.7cm} (c) RISE\hspace{1cm}(d) SIDU~(proposed) \hspace{1.5cm}
\vspace{-0.3cm}
\caption{An example of failure of saliency maps to capture entire object class 'clock'.}
\label{f11}
\end{figure*}

In recent years deep neural networks (DNN) have resulted in ground-breaking performance in solving many complex and long-running problems of artificial intelligence (AI). In particular, employing DNN architectures in tasks such as object detection~\cite{blend},  image classification~\cite{classy} and medical imaging~\cite{Gonz2020} has received great attention within the AI research field. As a result, it is no surprise to observe that DNNs have become a favoring solution for any applications involving big data analysis. As human dependency on these solutions increase on a daily basis, it is crucial from both research and business standpoints to understand the underlying processes of DNNs that output a certain decision.~As reported in recent works~\cite{Li21,bai2021explainable}, such decisions result from the complex inner stacked layer of the DNN that are typically referred to as 'black-box' model. The use of the term 'black-box' indicates how it is very challenging to understand which inner features of the model are the major contributors to the accuracy of the output~\cite{shin2021embodying}. In such cases the term 'black-box' predictors is used to aid such comprehension aspects. The interpretation ability of the 'black-box' DNN provides transparent explanation and audit model output that is crucial for sensitive domains such as medical or risk analysis ~\cite{shin2021people,shin2021effects}. Consequently, a new paradigm addressing explainability of these models has emerged in AI research namely \textbf{Explainable AI} (XAI)~\cite{weitz2020let}. XAI attempts to provide further insight into the black-box models and  their internal interactions that enable humans to understand a machine-generated output. Furthermore, for end-users in sensitive domains, XAI gives the ability to interpret model features at the 'group level' or 'instance level' of the input  which results in gaining greater trust  for validating the outcome of deployed AI models.
Although, there is no standard consensus in the literature regarding how to define a human-interpretable explanation method forthe black-box model, a widely-adopted and popular approach is to form a visual saliency map of input data showing which parts of the input have influence on the final prediction. This is motivated by the fact that the visual explanation methods can align closely with human intuition. For instance, it is more straightforward to the end-user in the medical domain to evaluate and compare the visual saliency map on a medical image produced by a DNNs model with those generated by actual clinicians. A number of visual explanation algorithms has been proposed among which methods such as LIME~\cite{ribeiro2016should}, GRAD-CAM~\cite{selvaraju2020grad} and RISE~\cite{Petsiuk2018rise} are the most used examples of this class. While each of these methods can be justifiable in one way or another, apart from challenges such as gradient computation of DNN architecture (e.g., Grad-CAM) or visualizing all the perturbations modes (e.g., RISE), the generated visual explanation suffers from a lack of localizing the entire salient regions of an object, which is often required for higher classification scores.
Following our prior identification of this research gap in the field, we further define it by proposing a new visual explanation approach known as SIDU~\cite{SIDU_IEEE} to address issues relating to salient region localization. SIDU stands for 'Similarity Difference and Uniqueness' method for estimating pixel saliency  by extracting the last convolutional layer of the deep CNN model and creating the similarity differences and uniqueness 
masks that are eventually combined to form a final map for generating the visual explanation for the prediction. We briefly showed by both quantitative and qualitative analysis how SIDU can provide greater trust for the end-user in sensitive domains. The algorithm provides improved localization of the object class being questioned (see, for example, Figure.~\ref{f11}~d).). 
This results in gaining greater trust of human expert level to rely on the deep model. This paper aims at providing a more general framework of the SIDU method by presenting the proposed method in further details whilst exploring its characteristic via various experimental studies. Concretely, the studies investigate SIDU's visual explanation through three main levels of evaluation as proposed in~\cite{doshi2018considerations}. Since these evaluation methods have different pros and cons, the superior performance of the SIDU can be investigated at depth to provide a deeper level of insight. To the best of our knowledge, our comprehensive experiment studies of these different evaluation levels are the first  in the context of XAI. Moreover, the ability of the XAI method to generalize its explanations of the black-box in different deployment scenarios can establish further trust. As evident in recent work, one example where black-box models are subject to less generalization is the presence of adversarial attack especially in sensitive domains and wider scope of trust~\cite{Ann2020}. Therefore, we investigate how XAI can handle such potential threat and respectively guard against it. Our \textbf{main contributions} in this work can be summarized  as follows:

\begin{enumerate}
    \item We provided step-by-step detailed explanations of the SIDU algorithm that from our investigation yielded a visual explanation map, which enabled localization of entire object classes from within an image of interest.
    \item We conducted three different types of experimental evaluations to thoroughly assess SIDU: these were coined as (1) 'Human-Grounded', (2) 'Functionally-Grounded', and (3) 'Application-Grounded' evaluations. Initially for (1) we conducted an interactive experiment with eye-tracking non-expert subjects to develop a database containing natural image annotation. This was done to assess how closely human eye-fixation on natural images can be matched to the visual salient map of SIDU to recognize the object class. In a similar setting, (3) was performed to assess the retinal quality assessment, and (2) was implemented alongside an automatic casual metrics~\cite{Petsiuk2018rise} on two datasets with different characteristics.
    \item Robustness of SIDU's explanation was analyzed in the presence of adversarial attacks to show how different noise levels can affect the classification task of the black-box model as well as its explanation consistency.
 
\end{enumerate}
The rest of the paper is organized as follows. Section~\ref{sec:related_work} presents state-of-the art XAI methods, XAI evaluations methods, and adversarial attacks. SIDU is explained in section~\ref{sec: proposedmedthod} with section~\ref{evl} having four subsections that are devoted to a particular evaluation of SIDU. In section~\ref{sec:functionally_grounded}, Functionally-Grounded evaluation is presented. In section~\ref{sec:human_grounded}, Human-Grounded evaluation is applied and 
Application-Grounded evaluation in section~\ref{sec:application grouded} is used to assess SIDU's performance. In section~\ref{sec:adv_noise_experiments}, evaluation of SIDU with respect to adversarial attack is shown and lastly section~\ref{sec:conclusion} concludes the study and discuss future work.
\section{Related Work} \label{sec:related_work}
In this work, we follow three main research directions of XAI: a) visual explanation methods developed to explain the black-box model such as deep CNN, b) validity and evaluation of the generated explanation by XAI methods  and c) vulnerability of black-box explanation method toward adversarial attacks. The literature of each direction is presented in the following subsections.
\subsection{Visual Explanation}
For an end-user, visual explanation methods makes it easier to understand the prediction output of the black-box model. One common approach to generate such a visualization is done via \textbf{saliency maps}~\cite{li2020enhanced,ren2013band} and such algorithms may be divided into the following \textit{three }categories:'back-propagation based' methods, 'perturbation-based' methods and 'approximation-based' methods. \textbf{Back-propagation methods}: back-propagation methods spread a feature signal from an output neuron rearwards through the layers of a model to the input in a single pass; making them efficient. 'Layer wise Relevance Propagation'~\cite{montavon2019layer} and 'DeCovNet' \cite{bach2015pixel} are examples of this category. Network weights and feature activation map of CNN model at a specific layer, e.g., CNN's last layer, are considered as an effective saliency method for generating visual explanation. Class Activation Mapping (CAM)\cite{zhou2016learning} that visually highlights the discriminative region of the image class prediction is an example of this family. In addition, the gradient or its modified version in the back-propagation algorithm can be employed to visualize the derivative of the CNN's output w.r.t. to its input, e.g. such as Grad-CAM \cite{selvaraju2020grad}. An improved method to produce input images that effectively activate a neuron was proposed in~\cite{simonyan2013deep}. The method explored in this related work was focused upon generating class-specific saliency maps by performing a gradient ascent in pixel space to reach a maxima. This synthesized image served as a class-specific visualization that augmented comprehension of how a given CNN modeled a class. \textbf{Perturbation-based methods}: here, the input is perturbed while keeping track of the resultant changes to the output. In some work, the change occurs at intermediate layers of the model. The state-of-the-art RISE \cite{Petsiuk2018rise} algorithm belongs to this category. Meaningful perturbations~\cite{fong2017interpretable} optimized a spatial perturbation mask that maximally effects a model’s output to reveal a new image saliency model that sought to identify where an algorithm searches by finding out which regions of an image most affected its output level when perturbed. \textbf{Approximation-based method}:
Methods of this class attempt to provide explanation to a complex black-box model by utilizing an easier-to-understand and more interpretable model such as decision trees or linear regression. 
Apart from these simple models, a good example class that is widely applied to visual input is the LIME algorithm~\cite{ribeiro2016should}. The main idea behind this related approach was to sample single visual input (i.e., image patches), correlate to the predictor model and subsequently identify its contribution toward the output class. The prediction results of each sample patch of the single image were then weighted with respect to the highest class score respectively. Finally, these weightings were used to train a simple surrogate model that was used as a local explanation for the result of the complex
model. Furthermore, another related work titled DeepLift~\cite{shrikumar2017learning}  evaluated the importance of each input neuron for a particular decision by approximating the instantaneous gradients (of the output with respect to the inputs) with discrete gradients. This obviated the need to train
interpretable classifiers for explaining each input-output relationship (as in LIME) for every test point.Inspired by the CAM method under the back-propagation based visual saliency approach, our proposed visual explanation, SIDU~\cite{SIDU_IEEE}  utilized 'Similarity Difference' and 'Uniqueness' measures to score the importance of associated activation maps from the last convolution layer of a CNN model. The proposed visual explanation algorithm is a gradient-free method that can effectively localize an entire salient region of the object of interest compared to the state-of-the-art XAI methods such as Grad-CAM and RISE.
\subsection{ Evaluation of Explanation Methods}
Since it is rather challenging to establish a unique and generalized evaluation metric that can be applied to any task, authors in ~\cite{doshi2018considerations} proposed  three different types of evaluations to measure the effectiveness of explanations. These are presented in the following.
\begin{enumerate}
\item \textbf{Application-Grounded evaluation}: Application-Grounded evaluation includes carrying out human experiments within a real application. If the researcher has a concrete application in mind—such as teaming up with doctors on diagnosing patients with a specific disease—the best method to show that the design is effective is to assess it with respect to the task.  A sound experimental setup and knowing how to evaluate the quality of the elucidation are needed. This approach is based upon how well a human can expound how the same (machine) decision is reached as output. Human expert level evaluation is necessary for those end-users who may have less confidence in the prediction model (e.g., clinician).
\item \textbf{Human-Grounded evaluation}: Human-Grounded evaluation  involves conducting basic human-subject experiments that ,substantiate the core of target application. This method is appealing when experiments involving the target community are difficult. The evaluations can be completed with laypersons, thus creating a greater subject pool and cutting down expenses, since we do not have to pay highly trained domain experts. 
\item \textbf{Functionally-Grounded evaluation}: This method utilizes numeric metrics or proxies such as 'local fidelity' to evaluate explanations across different applications. The main advantage of this evaluation is that it is free from human bias that effectively saves time and resources. Most of the state-of-art methods fall into this category~\cite{bach2015pixel,fong2017interpretable}. For example, the authors in~\cite{Petsiuk2018rise} proposed casual metrics \textit{insertion} and \textit{deletion}, which are independent of humans to evaluate the faithfulness of the XAI methods.
\end{enumerate}
 \subsection{Adversarial Attacks}
In the context of XAI, adversarial attack generators can be divided into 'white-box' attacks and 'black- box' attacks. The Fast Gradient Sign Method (FGSM)~\cite{goodfellow} and Projected Gradient Descent (PGD)~\cite{pgd} algorithms are well-known examples of a white-box attack where small amount of noise is added to an image that is not visually detectable by the end user. In the case of black-box attacks, the adversarial attack happens through various mechanisms to fool the model's classifier and alter its outcome. The majority of the proposed approaches in this class are based on perturbing the model input either globally or locally. For instance, DeepFool~\cite{deepfool} attack can be characterized by performing pixel-wise perturbation of an image while an adversarial patch attempts to change the pixel values in a specific region of an image. In general, the ability of changing a model's output via small input perturbations makes the XAI explanation methods challenging and less reliable. Thus, to establish greater trust, it is essential for the XAI algorithms to only be effective but also robust against an adversarial attack at the same time~\cite{Ann2020}. Analyzing how the black-box explanation ( like SIDU ) can effectively handle such a potential problem helps the end-user to guard against a possible disastrous outcome from the classifier when adversarial attack is presented.

\begin{figure}[t!]
\begin{center}
 \includegraphics[width=\textwidth,  height = 7cm]{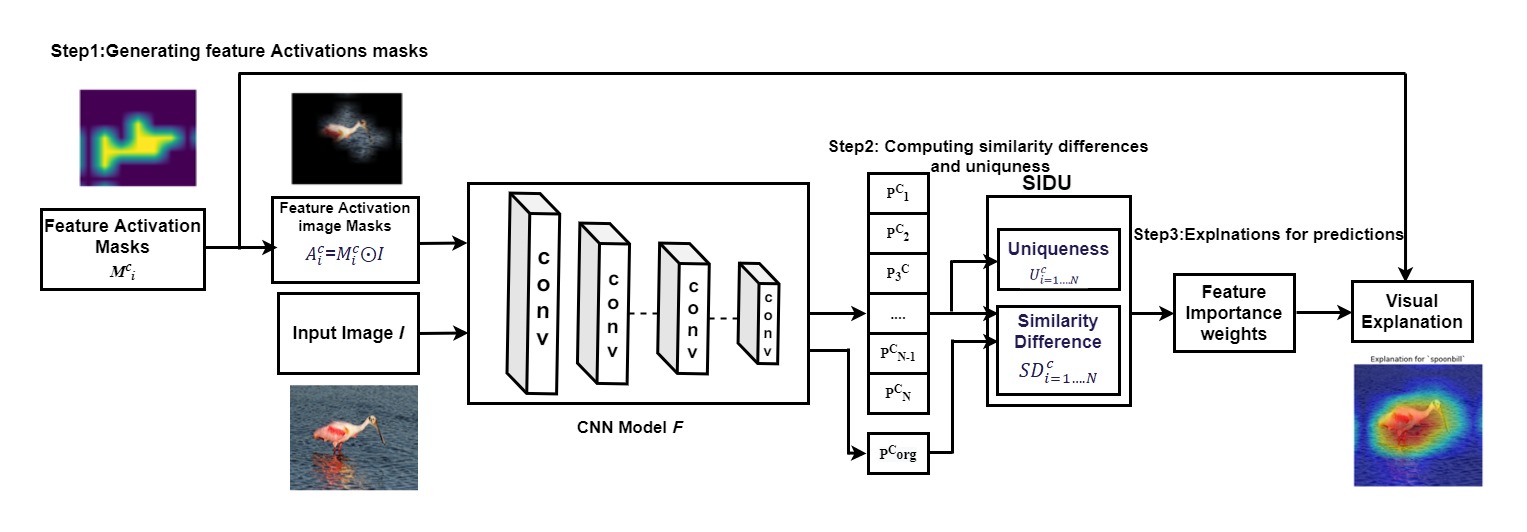}
 \vspace{-1.0cm}
\caption{Block diagram of SIDU. The CNN model $F$ is same of all the steps.}
\label{fig:sidu_block_diagram}
\end{center}
 \end{figure}
\vspace{-.3cm}

\section{SIDU: Proposed Method} \label{sec: proposedmedthod}
Recent XAI methods have shown that deeper representations in CNN models illustrate higher-level visual features~\cite{bai2021explainable}.
A recent approach titled as Grad-CAM~\cite{selvaraju2020grad} interprets the importance of each neuron responsible for a decision of interest by computing the gradient information from the last convolutional layer of the CNN. Alternatively, the authors in~\cite{Petsiuk2018rise} proposed a method titled RISE, which finds the effect of selectively inserting or deleting parts of the input (\textit{perturbation-based}) in the CNN model's output prediction. This perturbation-based method has been found to provide increased accuracy of visual explanation saliency maps compared to gradient based methods, However these methods fail to visualize all the perturbations in order to determine which one characterizes the best desired explanation. Furthermore, the visual explanations generated by both the gradient-based and perturbation explanations methods failed to localize the entire salient regions of an object class responsible for higher classification scores.

To overcome the challenges of the most recent state-of-the art methods we proposed a XAI method that consequently provides better explanation method for any given CNN model. The proposed method takes the last convolution layer for generating the masks. From these masks Similarity Difference and Uniqueness scores are computed to get the explanation of the CNN model decision acronymed in therefore denoted SIDU. An overview of the proposed method is presented in Figure~\ref{fig:sidu_block_diagram}. Our method is composed of three steps, First we extract the last convolution layer of the CNN to generate the feature image mask using the last convolution layer of the given model. Second, we compute the similarity differences for each mask with respect to a predicted class and finally we compute the weights of each mask and combine them into a final map that shows the explanation of the prediction. Each step is described in the following subsections~\ref{sec:generation_masks} $-$~\ref{sec: step3_visual_exp}. Note that, the CNN model used is the same for all steps. 

\vspace{-.3cm}

\begin{figure}[t!]
\begin{center}
 \includegraphics[width=\textwidth, height = 4.5cm]{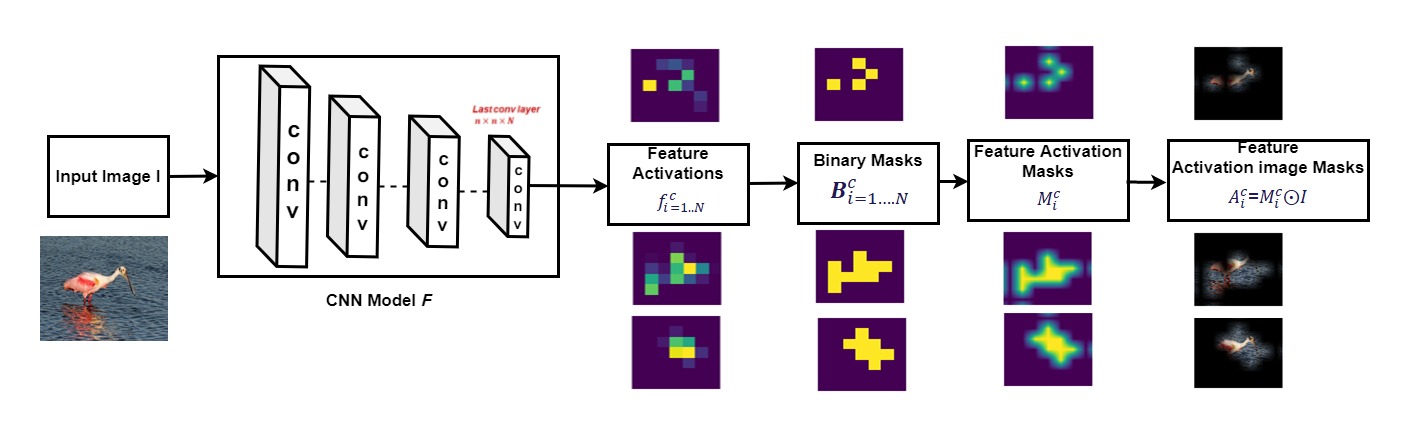}
 \vspace{-1.0cm}
\caption{The procedure of generating feature image masks from last layer activation's of CNN model $F$. The total numbers of masks generated are equal to the size of last convolution layer of CNN model $F$. We have shown some of the feature activation image masks  $A^{c}_{i= 1, 500}$ in the Figure. Note that the CNN model $\textit{F}$ used is same for all the steps.} \label{fig:generating masks}
\label{f22}
\end{center}
 \end{figure}
 
\subsection{Step1: Generating Feature Activation Image Masks} \label{sec:generation_masks}

To provide a visual explanation of the predicted output of a CNN model $\textit{F}$, we first generate  feature activation image masks from the last convolution layers. For any deep CNN model $\textit{F}$, we consider the last convolution layers of size $n\times n \times N$ where $'n'$ is the size of that convolution layer and $'N'$ is the total number of features activation $\textbf{f}$ of class $c$, i.e., $\textbf{f}^{c} = [f^{c}_{1},....f^{c}_{N}]$. For example, if the CNN model $\textit{F}$  has the last convolution layers of size $7\times 7\times 2042$, the total feature activations we can generate is $2042$ of size $7\times 7$. Therefore, the activation masks are generated upon image class explanation. Each feature activation map $f^{c}_{i}$ is then converted into a binary mask $B^{c}_{i}$ by thresholding each value and is given by
\begin{equation}
     {B^{c}_{i=1..N} = f^{c}_{i= 1..N} > \tau}
\end{equation}
where $ \tau$ is the threshold. In our experiments we use $\tau= 0.5$.~ Note that we found experimentally that choosing different threshold values in the mask binarization step has almost no effect on generating the final explanation heatmap of the input image.
 The binary mask $B^{c}_{i}$ is then up-sampled by applying bi-linear interpolation for a given input image $I$ with size of $Width \times Height$. Next, the up-samples binary mask $M^{c}_{i}$ will have values between $[0,1]$ and it is no longer binary. The up-sampled binary masks are also known as feature activation masks and is shown in Figure~\ref{fig:generating masks}. Finally, point-wise multiplication is performed between the feature activation mask (Up-sampled binary mask) $M^{c}_{i}$ and input image $I$ to calculate the feature activation image mask $A^{i}_{c}$ and is represented as
\begin{equation} 
  { A^{c}_{i} = F(I \odot M^{c}_{i}),}
\end{equation}
where $F$ is an CNN model, $A^{c}_{i}$ is the feature activation image mask of feature map $f^{c}_{i}$ and $i= 1,....N$. The procedure of generating feature activation image masks is shown in Figure~\ref{fig:generating masks} where we illustrate some of the feature activation image masks from the total number of masks $N$ . The feature activation image masks $A^{c}$ of object class $c$ are used to get prediction scores which is explained in detail in the following subsection~\ref{sec:simi_uniq_cal}
\vspace{-.25cm}

\subsection{Step2: Computing feature importance weights using Similarity Differences and Uniqueness} \label{sec:simi_uniq_cal}

\begin{figure*}[t!]
\begin{center}
 \includegraphics[width=\textwidth, height = 5cm]{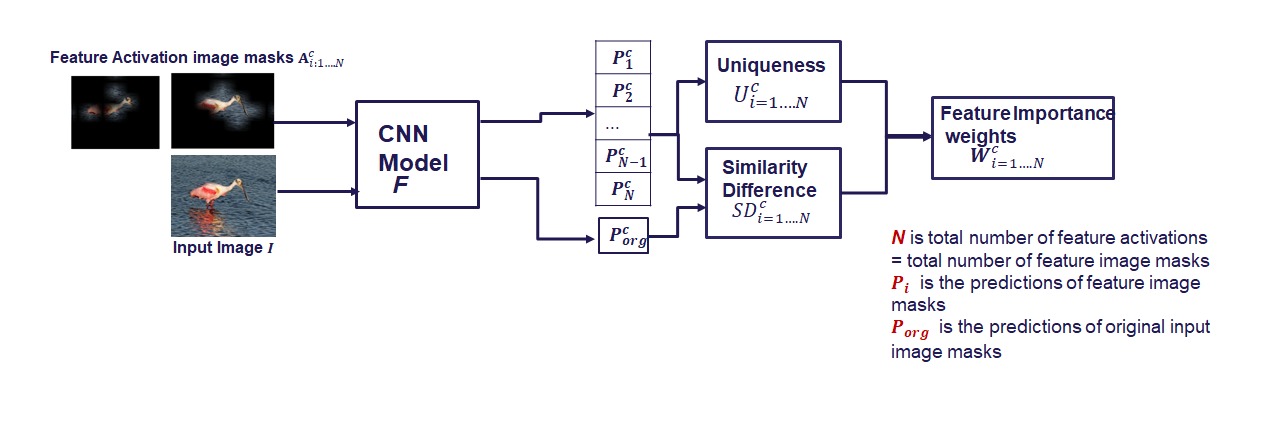}
 \vspace{-1.3cm}
\caption{The prediction score vectors for each individual feature  activation image mask $A^{c}_{i}$ and the original image $I$ are computed from the CNN model $F$. These prediction score vectors are used for computing Similarity Differences and Uniqueness, and finally the dot product is calculated to get the feature importance weights. Note that the CNN model $F$ is the same for all the steps.}\label{fig:prediction score vecotors for sidu}
\label{f23}
\end{center}
 \end{figure*}

The total number of feature activation image masks is dependent on the number of activations in the last convolution layer of the CNN model. Let the last convolution layer of the CNN model $F$ be of size $n \times n \times N $. The total number of feature activation image masks will be $N$. Next, we compute probability prediction scores for all the feature activation image masks $\textbf{A}^{c}$ of object class $c$, i.e., $\textbf{A}^{c}= [A^{c}_{1},....A^{c}_{N}]$ individually using the same CNN model $F$ used for generating the feature activation image masks. The probability prediction score of the feature activation image mask $A^{c}_{i}$ is defined as $P^{c}_{i}$ and the probability prediction score for the given input image $I$ is defined as $P^{c}_{org}$. The prediction scores vector size will be dependent on the total number of classes use to train the CNN model. E.g., If the CNN model is trained on the ImageNet dataset, which has a total of 1000 object classes, then the size of the predictions score vector ${P}^{c}_{i}$ of the each individual feature image mask $A^{c}_{i}$ will be $1 \times 1000$,  where $i= 1...N $. Figure~\ref{fig:prediction score vecotors for sidu} on page~\pageref{fig:prediction score vecotors for sidu}, illustrates the procedure of computing the predictions scores vector.

Once the predictions scores vectors are computed for all feature activation image masks and original input image, we then compute similarity differences between each input feature activation image mask prediction score $P^{c}_{i}$ and prediction score $P^{c}_{org}$ of the original input image $I$. The similarity difference between these two vectors gives the relevance of feature activation image mask with respect to the original input image. The intuition behind computing the relevance of a feature map is to measure how the prediction changes if the feature is not known, $i.e.,$ the similarity difference between prediction scores. The relevance value of the feature activation image mask will be high if it is similar to the predicted class but the relevance value will be low if dissimilar. The Similarity Difference measure between the prediction vector of the original input image $I$, $P^{c}_{org}$ and the $i^{th}$ feature activation image mask prediction, $P^{c}_{i}$ is given by
\begin{equation}
 {SD^{c}_{i} = \exp{(\frac{-\|P^{c}_{org}-P^{c}_{i}\|}{2\sigma^2})}}\label{e3}
\end{equation}
where $\sigma $ is an controlling parameter. It should be noted from Eq.\ref{e3} that ${P}^{c}_{i}$ is the prediction vectors for the feature activation image mask $A^{c}_{i}$ generated from the last convolution layer of CNN model $F$. This is illustrated in Figure.4.
Moreover, the Similarity measure in Eq.\ref{e3} is inspired by Gaussian kernel function which is a suitable metrics for weighting observations as opposed to Euclidean distance. The kernel function decreases with distance and lies between zero and one. For Euclidean distance, however, the value increases with distance and provides only an absolute difference between two vectors.
After computing the similarity difference measure, we also computed a uniqueness measure $U^{c}$ between the feature activation image masks prediction score vectors. 
It is one of the most popular assumptions that the image regions which stand out from the other regions grab our attention in certain aspects. Therefore the region should be labeled as a highly salient region. We therefore evaluate how different each respective feature mask is from all other feature masks constituting an image. The reason behind this is to suppress the false regions with low weights and highlight the actual regions which are responsible for predictionswith higher weights.
The uniqueness measure for the $i^{th}$  feature image mask of object class $c$, $U^{c}_{i}$, is defined as 
\begin{equation} \label{eq:uniqness}
    U^{c}_{i} = \sum_{j=1}^{N}\|P^{c}_{i}-P^{c}_{j}\|,{~~~~~~~i= 1,2,..., N}
\end{equation}
Where $N$ is the total number of feature activation image masks. Finally, the weight of each feature importance $W^{c}_{i}$ is computed as the dot product of the Similarity Difference $SD^{c}_{i}$ and Uniqueness measure $U^{c}_{i}$ where
\begin{equation} \label{eq:sim_uniqness}
  {W^{c}_{i} = SD^{c}_{i}\cdot U^{c}_{i}},
\end{equation}
where $SD^{c}_{i},U^{c}_{i}$ are the Similarity Difference and Uniqueness values for the feature activation image mask $A^{c}_{i}$ of the object class $c$. The total number of feature importance weights will be as size of total number of masks $N$. The feature importance weight will be high for the feature which has more influence in predicting the actual class object $c$ and low for the feature with low influence.

\subsection{Step3:Visual Explanations for the prediction} \label{sec: step3_visual_exp}
\begin{figure}[t!]
\begin{center}
 \includegraphics[width=\textwidth]{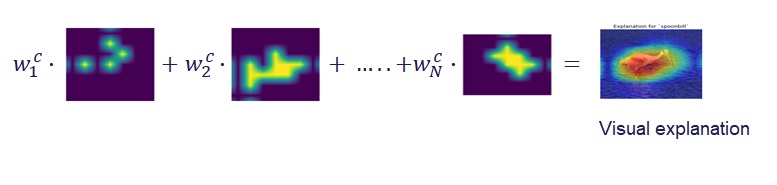}
 \vspace{-1.0cm}
\caption{Visual explanation for the prediction. The visual explanation is a weighted linear combinations of feature activation masks for the prediction of the class}\label{fig:vis}
\label{f23}
\end{center}
 \end{figure}

To get the visual explanation (saliency map) of the predicted output class $c$ of a CNN model $F$, we then performed a weighted sum between feature activation mask $M^{c}_{i}$ and the corresponding feature importance weights $W^{c}_{i}$,  where the weights are computed by Eq.~\ref{eq:sim_uniqness}. The visual explanation map is in the form of a heatmap (saliency map) and is represented as $S_{c}$ and is shown in Figure~\ref{fig:vis} on page~\pageref{fig:vis}. The visual explanation map $S_{c}$ is also known as the class discriminative localization map. Thus, the visual explanation of the predicted class $c$ is given by
\vspace{-.2cm}
\begin{equation} \label{eq: visual_exp}
    S_{c} = \frac{1}{N}\sum_{i=1}^{N} W^{c}_{i} \cdot M^{c}_{i}
\end{equation}
 The weighted combinations of feature activation masks to calculate the final visual explanation (saliency map) of the prediction of the class is illustrated in Figure~\ref{fig:vis}. 

\begin{figure*}[t!]
\centering

\includegraphics[width= 0.22\textwidth, height = 2.4cm]{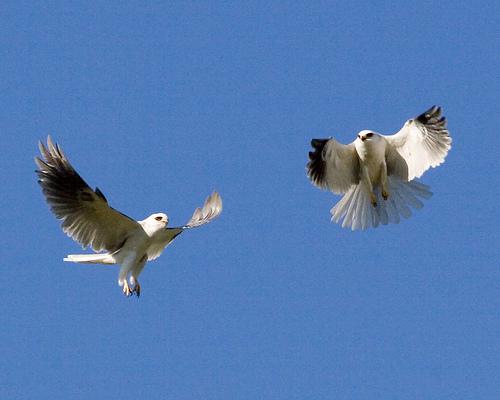}
\hspace{0.01\columnwidth}
\includegraphics[width= 0.22\textwidth, height = 2.4cm]{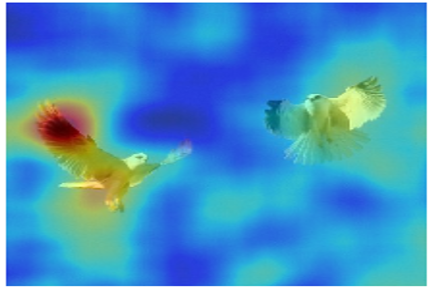}
\hspace{0.01\columnwidth}
\includegraphics[width= 0.22\textwidth, height = 2.4cm]{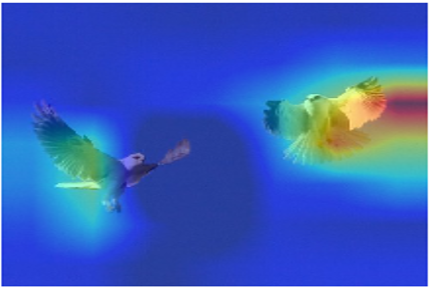}
\hspace{0.01\columnwidth}
\includegraphics[width= 0.22\textwidth, height = 2.4cm]{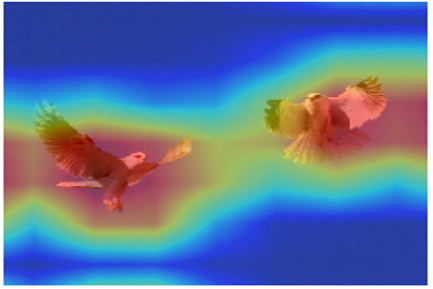}
\hspace{0.01\columnwidth}

\par

\centering
\vspace{1mm}
 
\par
\centering
\includegraphics[width= 0.22\textwidth, height = 2.4cm]{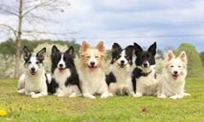}
\hspace{0.01\columnwidth}
\includegraphics[width= 0.22\textwidth, height = 2.4cm]{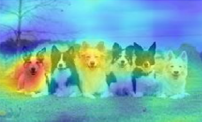}
\hspace{0.01\columnwidth}
\includegraphics[width= 0.22\textwidth, height = 2.4cm]{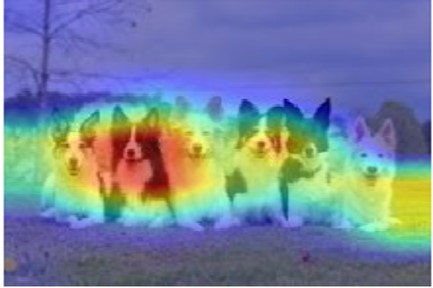}
\hspace{0.01\columnwidth}
\includegraphics[width= 0.22\textwidth, height = 2.4cm]{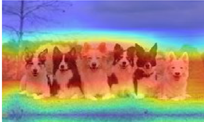}
\hspace{0.01\columnwidth}
\par
\centering
\includegraphics[width= 0.22\textwidth, height = 2.4cm]{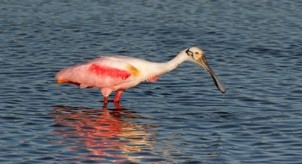}
\hspace{0.01\columnwidth}
\includegraphics[width= 0.22\textwidth, height = 2.4cm]{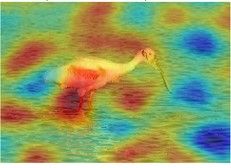}
\hspace{0.01\columnwidth}
\includegraphics[width= 0.22\textwidth, height = 2.4cm]{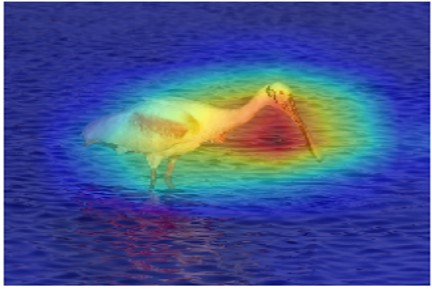}
\hspace{0.01\columnwidth}
\includegraphics[width= 0.22\textwidth, height = 2.4cm]{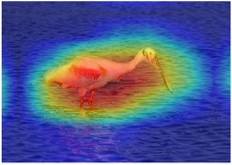}
\hspace{0.01\columnwidth}
\par
\centering
\vspace{1mm}
 
\par

\centering
\includegraphics[width= 0.22\textwidth, height = 2.4cm]{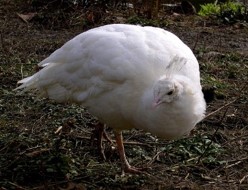}
\hspace{0.01\columnwidth}
\includegraphics[width= 0.22\textwidth, height = 2.4cm]{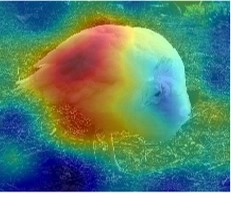}
\hspace{0.01\columnwidth}
\includegraphics[width= 0.22\textwidth, height = 2.4cm]{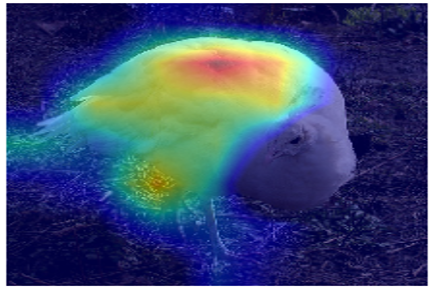}
\hspace{0.01\columnwidth}
\includegraphics[width= 0.22\textwidth, height = 2.4cm]{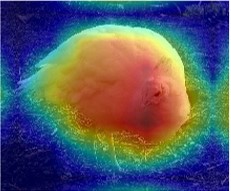}
\hspace{0.01\columnwidth}
\par
\centering
\vspace{1mm}

\par
\centering
\includegraphics[width= 0.22\textwidth, height = 2.4cm]{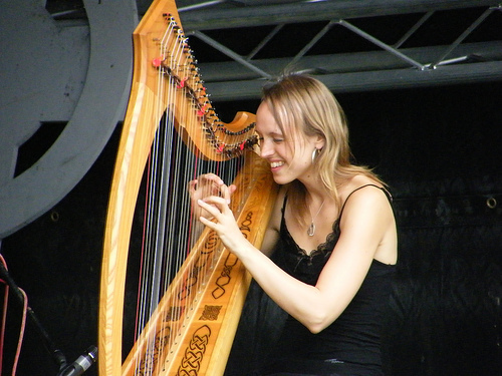}
\hspace{0.01\columnwidth}
\includegraphics[width= 0.22\textwidth, height = 2.4cm]{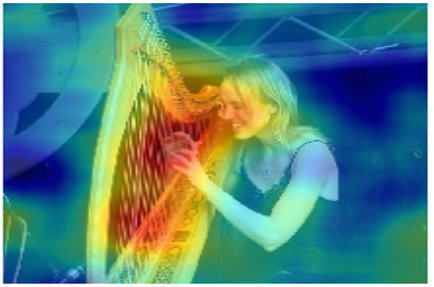}
\hspace{0.01\columnwidth}
\includegraphics[width= 0.22\textwidth, height = 2.4cm]{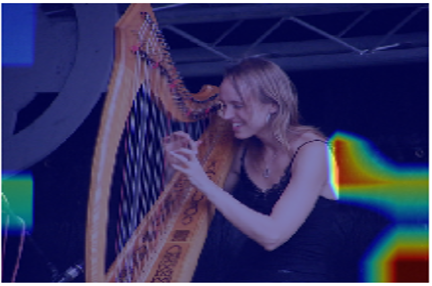}
\hspace{0.01\columnwidth}
\includegraphics[width= 0.22\textwidth, height = 2.4cm]{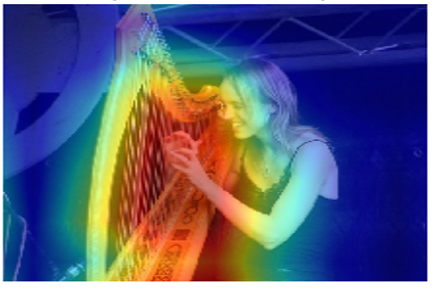}
\hspace{0.01\columnwidth}
\par
\centering
\vspace{1mm}
 
\vspace{1mm}
 
\par
\centering{}
\scriptsize(a) Original Image \hspace{1.1cm}(b) RISE\hspace{1.6cm}(c) GRAD-CAM \hspace{1.3cm} (d)~SIDU
\caption{Visual comparison of explanation maps generated for the natural images classes 'Bird','Borzoi dog', 'Spoonbill', 'Goose', and 'Harp'  predicted by CNN model.}
\label{f1s}
\end{figure*}

In summary, to explain the decision of the predicted class $c$ visually, we first generated the $N$ feature activation masks (up-sampled binary masks) from the last convolution layer of the deep CNN model $F$ which has $N$ number of feature activation maps of size $n \times n $. We then perform point wise multiplication between each generated up-sampled binary mask $M_{i}$ and the input image $I$ to calculate feature activation image mask. Next, we compute Similarity Differences $SD^{c}_{i}$ and uniqueness measure $U^{c}_{i}$ using predictions scores of feature activation image mask $A_{i}$. Feature importance weights $W_{i}$ of each feature activation image mask $A_{i}$ is computed by the dot product of $SD^{c}_{i}$ and  $U^{c}_{i}$. Finally, the visual explanation $S_{c}$ of a given input image is obtained by calculating a weighted sum of feature activation image masks $A_{i}$ as stated in Eq.~\ref{eq: visual_exp}. Furthermore an example of visual comparison of explanation maps generated for the natural images classes is illustrated in Figure.~\ref{f1s} on page~\pageref{f1s}.



\section{Evaluation}
\label{evl}

\noindent In this section we evaluate the performance of SIDU. We conducted a comprehensive set of experiments to study the correlation of the visual explanation with the model prediction to evaluate the faithfulness.
SIDU is evaluated using all three categories of evaluations as previously detailed herein~\cite{doshi2018considerations}, i.e., functionally grounded, application grounded, and human grounded. The evaluation results were compared with the most recent state-of-the art methods namely RISE~\cite{Petsiuk2018rise} and GRAD-CAM~\cite{selvaraju2020grad}. A good explanation method not only provides an appropriate explanation for the prediction but also it should be robust against adversarial noise. To this end, the proposed method is evaluated on adversarial samples and compared with the most recent state-of-the-art methods RISE~\cite{Petsiuk2018rise} and GRAD-CAM~\cite{selvaraju2020grad}. The experimental evaluation of faithfulness of the SIDU model on the above mentioned evaluation categories and effect of adversarial noise are described in section~\ref{sec:functionally_grounded},~\ref{sec:human_grounded},~\ref{sec:application grouded} and~\ref{sec:adv_noise_experiments}, respectively.

\subsection{Functionally-Grounded evaluation} \label{sec:functionally_grounded}

\begin{figure*}[hbt!p]
\begin{centering}
\includegraphics[width= 0.15\textwidth, height = 2.1cm]{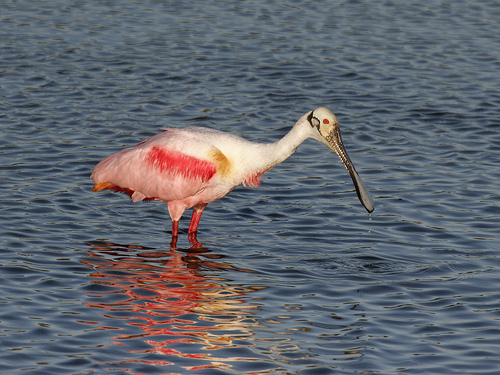}
\hspace{0.01\columnwidth}
\includegraphics[width= 0.15\textwidth, height = 2.1cm]{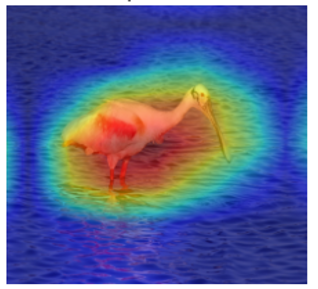}
\hspace{0.01\columnwidth}
\includegraphics[width= 0.15\textwidth, height = 2.3cm]{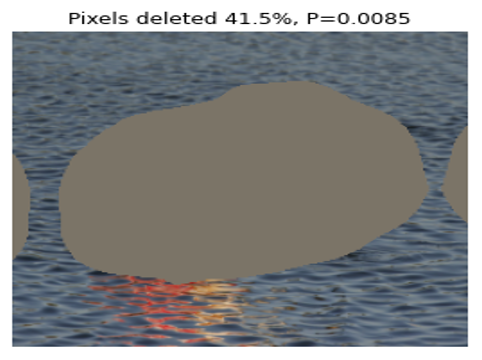}
\hspace{0.01\columnwidth}
\includegraphics[width= 0.15\textwidth, height = 2.3cm]{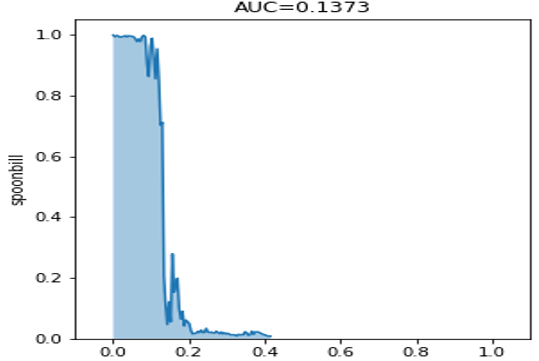}
\hspace{0.01\columnwidth}
\includegraphics[width= 0.15\textwidth, height = 2.3cm]{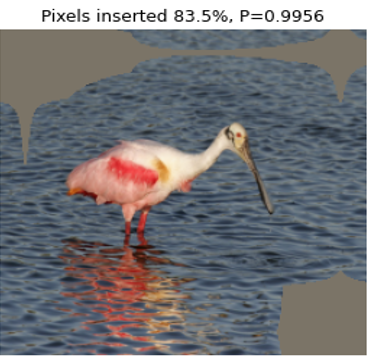}
\hspace{0.01\columnwidth}
\includegraphics[width= 0.15\textwidth, height = 2.3cm]{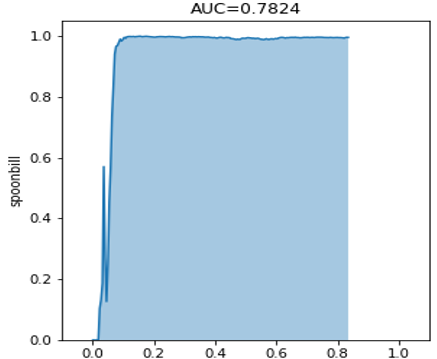}
\hspace{0.01\columnwidth}
\par\end{centering}\vspace{-.5cm}
~~~~~~(a) \hspace{1.5cm}(b) \hspace{1.5cm} (c)~\hspace{1.5cm}(d)~\hspace{1.5cm}(e)~\hspace{1.5cm}(f)~

\caption{Evaluation using \textit{insertion} and \textit{deletion} casual metrics AUC is computed. (a) original image (b) SIDU explanation map (c) the \textit{deletion} metric; this being where the salient pixels are gradually removed from the image for decreasing the importance,and the probability of the class 'spoonbill' as predicted by the CNN model is plotted with respect to the removed pixels Area Under Curve (AUC) is computed in (d). (e)  \textit{insertion} metric; this being where the salient pixels are gradually inserted to the image for increasing the importance, and the probability of the class 'spoonbill' predicted by the CNN model is plotted with respect to the inserted pixels and AUC is computed in (f)} \label{fig:inserion_deletion_metric}
\end{figure*}

To preform the Functionally-Grounded evaluation we choose the two automatic causal metrics \textit{insertion} and \textit{deletion} as proposed by~\cite{Petsiuk2018rise}. 
The \textit{deletion metric} deletes the saliency region in the image which is responsible for higher classification scores and forces the CNN model to change its decision. This metric estimates the decrease in the probability classification scores, when more pixels are removed from the saliency region. With the \textit{deletion} metric, the good explanation shows a sharp drop in the predicted score and area under the probability curve will be lower. Whereas, the \textit{insertion metric} measures the probability increase of the predicted score. As more pixels are inserted in the image, a higher Area Under Curve (AUC) rate can be achieved (i.e., effectiveness of explanation model at a greater level). The procedure of computing AUC using \textit{insertion} and \textit{deletion} is illustrated in Figure.~\ref{fig:inserion_deletion_metric} on page~\pageref{fig:inserion_deletion_metric}.
These metrics were selected since they are independent of human subjects, bias free and hence increase transparency when evaluating the XAI methods. 

In order to evaluate the performance of the SIDU explanation method we choose two datasets with different characteristics, namely- The ImageNet~\cite{ILSVRC15} dataset of Natural Images with 1000 classes. We used 2000 images randomly collected from the ImageNet validation dataset. The other is a  Retinal Fundus Image Quality Assessment (RFIQA) dataset from the medical domain consisting of 9,945 images with two levels of quality, 'Good' and 'Bad'. The retinal images were collected from a large number of patients with retinal diseases~\cite{SATYA_VISAPP}.
\noindent We conducted two experiments for evaluating the faithfulness of the proposed explanation method. The first experiment is performed on the ImageNet validation dataset where we randomly selected 2000 images from the ImageNet dataset. To do a fair evaluation, we choose two existing standard CNN models, ResNet-50~\cite{Resnet-50} and VGG-16~\cite{vgg16} that had been pre-trained on the ImageNet dataset~\cite{ILSVRC15}. Table \ref{tab:t_resenet_vgg16_merged} summarizes the results obtained on ResNet-50 for the proposed method and compares it to the most recent works RISE~\cite{Petsiuk2018rise} and GRAD-CAM~\cite{selvaraju2020grad}. It was observed that the proposed method achieved improved performance for both metrics, followed by RISE~\cite{Petsiuk2018rise} and GRAD-CAM~\cite{selvaraju2020grad}.  Table \ref{tab:t_resenet_vgg16_merged} summarizes the results obtained on the VGG-16 model for the proposed method and compares it to most recent works RISE~\cite{Petsiuk2018rise} and GRAD-CAM~\cite{selvaraju2020grad} where it can be identified that proposed method, SIDU achieved best performance. From the Table~\ref{tab:t_resenet_vgg16_merged}, we can observe that the values are better for ResNet-50 than VGG-16 for all the XAI methods, which suggests that ResNet-50 is a better classification model than VGG-16. Qualitative examples are shown in Figure.~\ref{f1s}. In our proposed method, the generated masks come from the last feature activation maps of the CNN model, due to this the final explanation map will localize the entire region of interest (object class).



\begin{table}[tb!]
\small
\caption{Comparision of XAI methods using ResNet-50 and VGG-16 on ImageNet validation set. All values in the table has the unit of Area Under Curve (AUC).}\label{tab:t_resenet_vgg16_merged}
\begin{centering}
\begin{tabular}{|
>{\columncolor[HTML]{C0C0C0}}c |c|c|c|c|}
\hline
\cellcolor[HTML]{C0C0C0} & \multicolumn{2}{c|}{Resnet-50~\cite{Resnet-50}} & \multicolumn{2}{c|}{VGG-16~\cite{vgg16}}                  \\ \cline{2-5} 
\multirow{-2}{*}{\cellcolor[HTML]{C0C0C0}\begin{tabular}[c]{@{}c@{}}XAI\\Methods \\ \end{tabular}} & \multicolumn{1}{l|}{Insertion$\uparrow$} & \multicolumn{1}{l|}{Deletion$\downarrow$} & \multicolumn{1}{l|}{Insertion$\uparrow$} & \multicolumn{1}{l|}{Deletion$\downarrow$} \\ \hline
RISE~\cite{Petsiuk2018rise}     & 0.63571 & 0.13505  & 0.47113 & 0.1313 \\ \hline
GRAD-CAM~\cite{selvaraju2020grad} & O.62863 & 0.15399  & 0.41720 & 0.15486 \\ \hline
SIDU     &  \textbf{0.65801} & \textbf{0.13424} & \textbf{0.49419} & \textbf{0.1309} \\ \hline
\end{tabular}
\par\end{centering}
\end{table}

\begin{table} [t!]
\small
\caption{Comparison of XAI methods on RFIQA dataset using trained ResNet-50 model. }\label{t2_rfiqa}
\begin{centering}
\begin{tabular}{|c|c|c|}
\hline 
METHODS & Insertion$\uparrow$ & Deletion$\downarrow$  \tabularnewline
\hline 
\hline 
RISE~\cite{Petsiuk2018rise} & 0.75231 & 0.59632 \tabularnewline
\hline
GRAD-CAM~\cite{selvaraju2020grad}& \textbf{0.91303} & \textbf{0.43061}   \tabularnewline
\hline 
SIDU &  0.87883 & 0.47818   \tabularnewline
\hline
\end{tabular}
\par\end{centering}
\end{table}
\noindent  We also conducted a second experiment on the Medical Image dataset which has totally different characteristics. We trained the existing ResNet-50~\cite{Resnet-50} with an additional two FC layers and softmax layer on the RFIQA dataset~\cite{SATYA_VISAPP}. The CNN model achieve $94 \%$ accuaracy. The proposed explanation method uses the trained model for explaining the prediction of the RFIQA  test subset with 1028 images. The evaluated results of the proposed method and RISE~\cite{Petsiuk2018rise} and GRAD-CAM~\cite{selvaraju2020grad} are summarized in Table~\ref{t2_rfiqa}. We can observe that the GRAD-CAM achieves slightly higher AUC for \textit{insertion} and lower AUC for \textit{deletion} followed by SIDU. RISE~\cite{Petsiuk2018rise} has shown least performance in both metrics, This can be explained by the fact that the RISE method generates $N$ number of random masks and the weights predicted for these masks give higher weights to false regions which makes the final map of RISE  noisy. The visual explanations of the proposed method (SIDU) and the RISE~\cite{Petsiuk2018rise}, GRAD-CAM~\cite{selvaraju2020grad} methods on the RFIQA test dataset are shown in Figure.~\ref{fig:f1s_medical}~(b),~(c),~(d). 

\vspace{-.3cm}

\subsection{Human-Grounded evaluation} \label{sec:human_grounded}
Human-Grounded evaluation is most appropriate when one aims at testing a general notions of an explanation quality. Therefore, for generic applications in the AI domain, such as object detection and object recognition, it might be sufficient to inspect a degree to which a non-expert human can understand the cause of a decision generated by a black-box model. 
One excellent way to measure and compare the correlation of visual explanation between a human subject and the black-box is to use an eye tracker that records the non-expert subject’s fixations within interactive test settings. This approach is chosen because of its similarity to XAI methods, visual explanations. Both generate heatmaps representing salient areas of an object in an image. 

An eye-tracker was used for gathering eye tracking data from human subjects to gain an understanding of visual perception~\cite{eye_jiang2014saliency}. The study using eye tracking data for understanding  human visual attention is useful and has received great attention by UX researchers~\cite{bergstrom2014eye}. For example, the authors in~\cite{eye_das2017human} conducted an experimental study and gathered data ‘human attention’ in Visual Question Answering (VQA) to interpret where the humans choose to look to answer the questions regarding the images. The authors in~\cite{eye_jiang2015salicon} established mouse-tracking approach to accurately collecting attention maps via collecting a large-amount of attention annotations for MS COCO on Amazon Mechanical Turk (AMT). In~\cite{mcdonnell2009eye}, recordings of subjects' eye-fixations in relation to body parts were used to investigate which body parts of virtual characters are most looked at in scenes containing duplicate characters or clones. However, all these experimental studies have used eye tracking to understand the human visual attention for different types of problems.

In our study, we investigated how non-expert subjects generated explanations via the eye-tracker, compared with those of generated by XAI visual explanation methods across natural images for recognizing object class. To this end, we follow the data collection protocol discussed in detail in the next section~\ref{sec:eye_tacking_data}.   

\subsubsection{Database of eye tracking data} \label{sec:eye_tacking_data}
We randomly sampled 100 images from 10 different classes of the ImageNet~\cite{ILSVRC15} benchmark validation dataset. All the collected images are RGB and are resized to $224 \times224$ pixels. 

\subsubsection{Data collection protocol}
\begin{figure*}[hbt!p]
\begin{centering}
\includegraphics[width= 0.23\textwidth, height = 2.3cm]{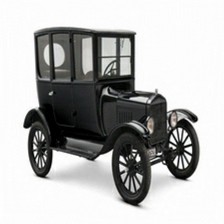}
\hspace{0.01\columnwidth}
\includegraphics[width= 0.23\textwidth, height = 2.3cm]{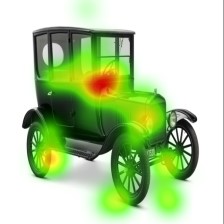}
\hspace{0.01\columnwidth}
\includegraphics[width= 0.23\textwidth, height = 2.3cm]{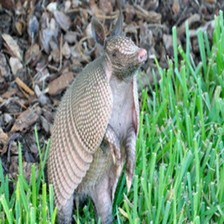}
\hspace{0.01\columnwidth}
\includegraphics[width= 0.23\textwidth, height = 2.3cm]{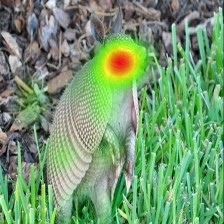}
\hspace{0.01\columnwidth}

\par\end{centering}
\caption{Examples of Eye-tracking data collection from humans for recognizing the given object classes 'Model T and 'Armadillo' }\label{fig:eye_tracking_images}
\end{figure*}

In order to collect eye-fixation, 5 human subjects participated in an interactive test procedure using Tobii-X120 eye-tracker in the following main steps:
\begin{enumerate}
\item The subject was seated in front of a computer-sized screen where the eye-tracker is ready to record the visual fixations and the system is calibrated.
\item Each image from the dataset was shown in a random order for 3 seconds and corresponding fixations of the subject were recorded.
\item We divided all 100 images into 4 equally sized data blocks with a break between each experiment in order to reduce the burden on each subject. We further add a cross-fixation image between two stimuli to reset the visionary fixation on the screen while changing from one image to the next.
\item The participants were shown random images from the collected dataset and then asked the question, what kind of object class is presented in the image.
\item The eye-fixations of each individual participant were automatically recorded via the eye- tracker when the participant looks at the image for recognizing the object class.
\item After all 5 participants' fixations were collected, an aggregated heatmaps was generated by convolving a
Gaussian filter across each user's fixation for each image-~see, Figure.~\ref{fig:eye_tracking_images} on page~\pageref{fig:eye_tracking_images}. The resulting heatmaps highlight the salient regions of each object class that often attracted attention of all subjects in the experiment and hence can be used to compare with the heatmaps produced by the XAI explanation algorithms.
\end{enumerate}

\subsubsection{Comparison Metrics} \label{sec:metrics}
To evaluate the models with human fixations using only one metric is not enough to achieve a valid and reliable outcome~\cite{riche2013saliency}. We used three metrics to compare the XAI and eye-tracker generated heatmaps~\cite{different_evalautaion_metrics}: These are (1) Area Under ROC Curve (AUC), (2) Kullback-Leibler Divergence (KL) and (3) Spearmans Correlation Coefficient metric (SCC) metrics. The use of multiple metrics ensures that the discussion about the results is as independent as possible from the choice of metrics. The results of the different evaluation measures are not necessarily the same, but when two metrics show similarities, then claims of robustness can be argued from a stronger position.
 \begin{enumerate}
 
\item\textbf{Area under ROC Curve (AUC):} \label{sec:auc}
The Receiver Operating Characteristics (ROC) is one of the commonly used metric for assessing the degree of similarity of two saliency maps. It is represented in the form of a graphical plot which describes the trade-off between true and false positives at different thresholds~\cite{different_evalautaion_metrics}. A fraction of true positives from the total actual positives are plotted against the false positives' fraction out of the total actual negatives to create the ROC. This is denoted as TPR, representing the true positive rate, and FPR that indicates the false positive rate. The rates are examined at different threshold values.
If a TPR value of 1 is achieved at 0 FPR, the prediction method is good. These values will yield a point in the ROC space's upper left corner and correspond to a near-perfect classification. Conversely, when the guess is completely random, it will generate a point along a diagonal line starting at the left bottom and going up towards the top right corner. If the diagonal divides the ROC space while and points above the diagonal, this represents good classification results. Such results are considered better than random results. On the other hand, the line below is a sign of poor results, which is even worse than getting random results. The Area Under Curve (AUC) is the method used to measure the ROC curve's performance. The AUC is equal to the probability of a classifier ranking a randomly selected positive instance, which is usually higher than a randomly selected negative instance, assuming that the positive ranks higher than a negative. To compute the AUC, XAI visual explanation heatmaps are treated as fixations' binary classifiers at numerous threshold values or value sets. The true and false positive rates are measured under each binary classified or level set to sweep out the ROC curve.

\item\textbf{Kullback-Leibler Divergence (KL-DIV):}\label{sec:kl-div}
The Kullback-Leibler Divergence is an metric, which is used to measure dissimilarity between two
probability density functions~\cite{different_evalautaion_metrics}.
For evaluating the XAI methods, eye-fixation maps and the visual explanation maps produced by the model are used for the distributions. $FM$ represents the heatmaps probability distribution from eye-tracking data, and $EM$ indicates the visual explanation maps probability distribution. These probability distributions are normalized and they are given by :

\begin{equation}\label{eq:kl_div_1}
    EM(x) = \frac{EM(x)}{\sum_{x=1}^{X}EM(x) + \epsilon},
    \vspace{0.5cm}
\end{equation}
\begin{equation}\label{eq:kl_div_2}
    FM(x) = \frac{FM(x)}{\sum_{x=1}^{X}FM(x) + \epsilon},
\end{equation}

where X is the number of pixels and $\epsilon$ is a regularization constant to avoid division by zero. The KL-DIV measure is computed between these two distributions to know whether the visual explanation map which is computed from the XAI method matches human fixations. It is a non-linear measure and generally varies in ranges from zero to infinity. If the KL-DIV measure between $EM$ and $FM$ is lower, then the $EM$ maps have better approximation of the human eye-fixation $FM$.
\item\textbf{Spearmans Correlation Coefficient (SCC):}\label{sec:scc}
Spearman’s correlation is a non-parametric measure that analyses how well the relationship between two variables can be described using a monotonic function~\cite{daniel1990applied}. It is a statistical method used mainly for measuring the correlation or dependency between two variables. This metric varies between the values of $- 1$ and $1$, where a score of $-1$, represents no correlation. The SCC between two variables will be high when observations have a similar ( with a correlation close to $1$) rank between the two variables, and low when observations have a dissimilar rank (with a correlation close to $- 1$) between the two variables~\cite{daniel1990applied}.

It is an appropriate measure for both continuous and discrete ordinal variables~\cite{daniel1990applied}. $FM$ represents the heat map from eye tracking data, whereas $EM$ is the visual explanation map. The SCC between the two random variable maps, $FM$ and $EM$ is given by :

\begin{equation}
    SCC (EM,FM) = \frac{cov(EM, FM)}{\sigma(EM) \times \sigma (FM)},
\end{equation}
\vspace{0.2cm}
where $cov(EM,FM)$ is the covariance of $EM$ and $FM$, $\sigma (EM)$ and $\sigma(FM)$ are the standard deviations of $EM$ and $FM$ respectively.

\end{enumerate}

\subsubsection{Comparing SIDU and State-of-art methods with human attention for recognizing the object classes}
 \label{sec:restuls}

\begin{figure*}[t!]

\begin{centering}
 
\par\end{centering}
\begin{centering}

\includegraphics[width= 0.18\textwidth, height = 2.5cm]{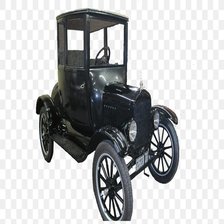}
\hspace{0.01\columnwidth}
\includegraphics[width= 0.18\textwidth, height = 2.5cm]{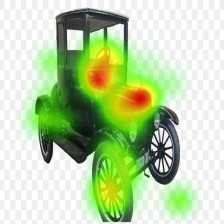}
\hspace{0.01\columnwidth}
\includegraphics[width= 0.18\textwidth, height = 2.5cm]{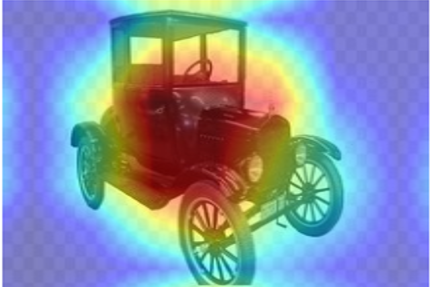}
\hspace{0.01\columnwidth}
\includegraphics[width= 0.18\textwidth, height = 2.5cm]{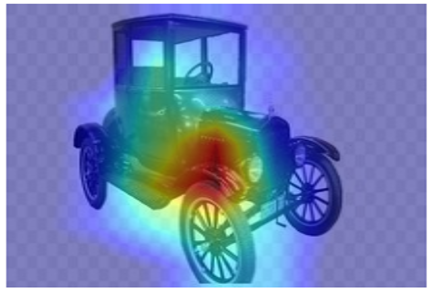}
\hspace{0.01\columnwidth}
\includegraphics[width= 0.18\textwidth, height = 2.5cm]{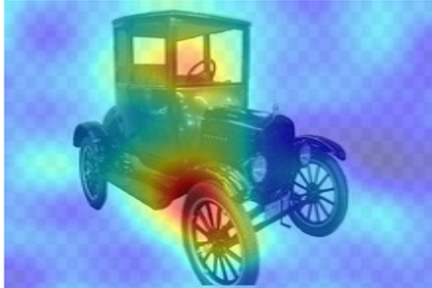}
\par\end{centering}
\begin{centering}
 
\par\end{centering}
\begin{centering}

\includegraphics[width= 0.18\textwidth, height = 2.5cm]{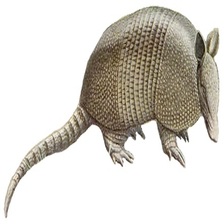}
\hspace{0.01\columnwidth}
\includegraphics[width= 0.18\textwidth, height = 2.5cm]{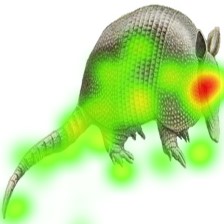}
\hspace{0.01\columnwidth}
\includegraphics[width= 0.18\textwidth, height = 2.5cm]{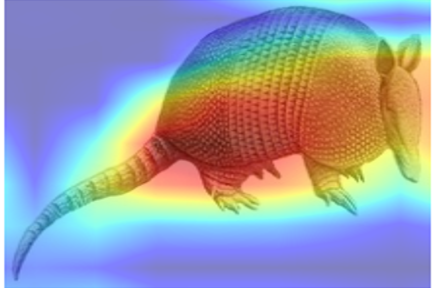}
\hspace{0.01\columnwidth}
\includegraphics[width= 0.18\textwidth, height = 2.5cm]{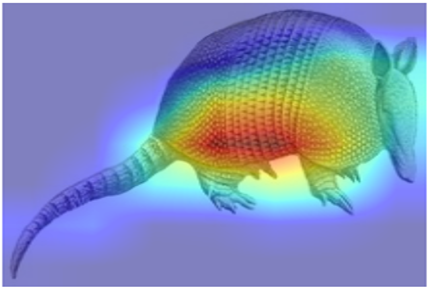}
\hspace{0.01\columnwidth}
\includegraphics[width= 0.18\textwidth, height = 2.5cm]{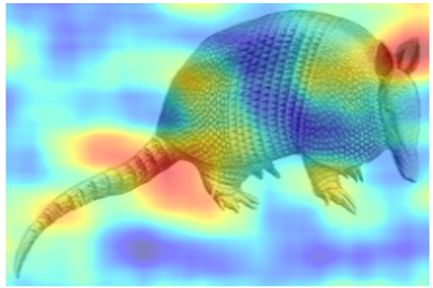}
\par\end{centering}
\begin{centering}
 
\par\end{centering}
\begin{centering}

\includegraphics[width= 0.18\textwidth, height = 2.5cm]{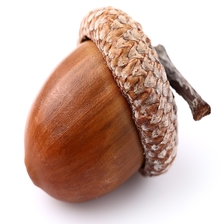}
\hspace{0.01\columnwidth}
\includegraphics[width= 0.18\textwidth, height = 2.5cm]{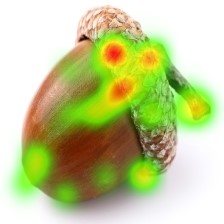}
\hspace{0.01\columnwidth}
\includegraphics[width= 0.18\textwidth, height = 2.5cm]{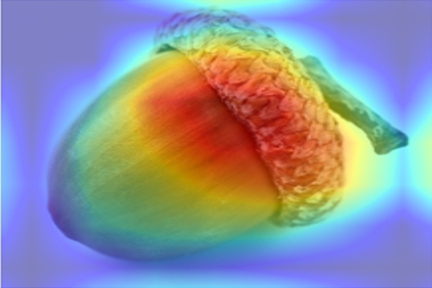}
\hspace{0.01\columnwidth}
\includegraphics[width= 0.18\textwidth, height = 2.5cm]{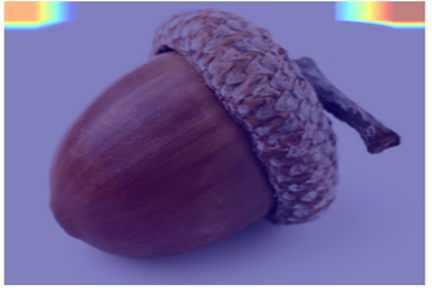}
\hspace{0.01\columnwidth}
\includegraphics[width= 0.18\textwidth, height = 2.5cm]{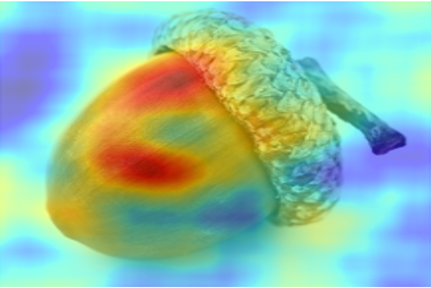}
\par\end{centering}
\begin{centering}
 
\par\end{centering}

\begin{centering}
 
\par\end{centering}
\begin{centering}

\includegraphics[width= 0.18\textwidth, height = 2.5cm]{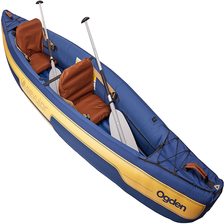}
\hspace{0.01\columnwidth}
\includegraphics[width= 0.18\textwidth, height = 2.5cm]{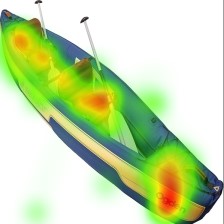}
\hspace{0.01\columnwidth}
\includegraphics[width= 0.18\textwidth, height = 2.5cm]{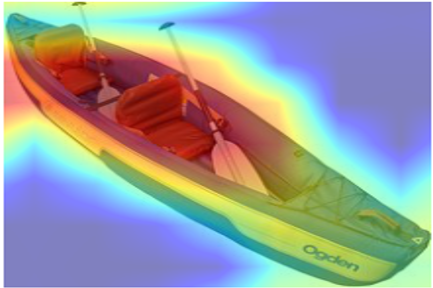}
\hspace{0.01\columnwidth}
\includegraphics[width= 0.18\textwidth, height = 2.5cm]{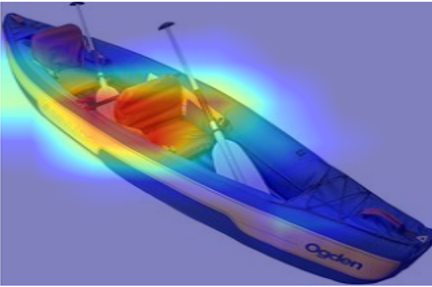}
\hspace{0.01\columnwidth}
\includegraphics[width= 0.18\textwidth, height = 2.5cm]{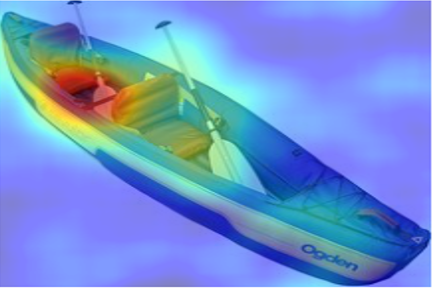}
\par\end{centering}
\begin{centering}
 
\par\end{centering}
\begin{centering}

\includegraphics[width= 0.18\textwidth, height = 2.5cm]{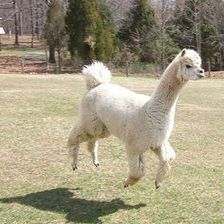}
\hspace{0.01\columnwidth}
\includegraphics[width= 0.18\textwidth, height = 2.5cm]{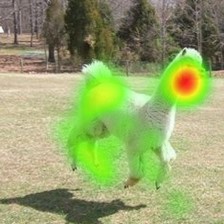}
\hspace{0.01\columnwidth}
\includegraphics[width= 0.18\textwidth, height = 2.5cm]{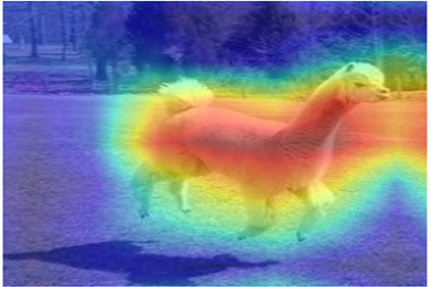}
\hspace{0.01\columnwidth}
\includegraphics[width= 0.18\textwidth, height = 2.5cm]{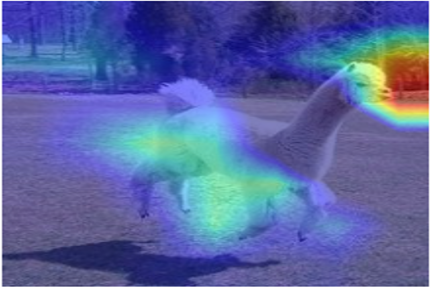}
\hspace{0.01\columnwidth}
\includegraphics[width= 0.18\textwidth, height = 2.5cm]{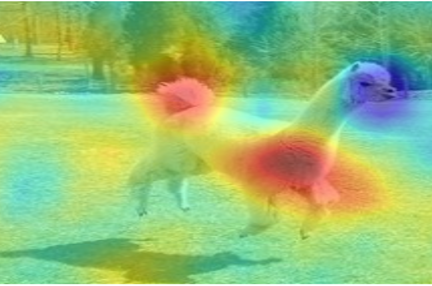}
\par\end{centering}
\centering{}
\scriptsize(a) Original Image\hspace{0.8cm}(b) Eye-tracker\hspace{0.8cm} (c)~SIDU\hspace{0.9cm}(d)~GRAD-CAM\hspace{0.8cm}(e)~RISE
\caption{Comparison of XAI methods visual explanation of object classes from top to bottom 'model T', 'armadillo', 'acorn', 'canoe' and 'kuvasz' with human visual explanation (heatmaps). The generated heatmaps in $3^{rd}$, $4^{th}$ and $5^{th}$  columns by the SIDU, GRAD-CAM and RISE demonstrate how the visual explanation methods are closely aligned with of human.  
} \label{fig:comparisions of visual explanation with experts}
\end{figure*}

In this experiment, we use the  Imagenet images eye-tracking data recordings described in section~\ref{sec:eye_tacking_data} to generate and evaluate the explanation by the XAI algorithms. To this end, we first generate ground truth heatmaps by applying Gaussian distributions on human expert eye-fixations. These heatmaps are then used to compare with the XAI  heatmaps. AUC, SCC and KL-DIV evaluation metrics are used to evaluate the performance. We finally calculate the mean of AUC, SCC and KL-DIV of all the images in the dataset. Table~\ref{tab:eye_tracking_comparisions} summarizes the results obtained by SIDU and the two different state-of-the art XAI methods RISE~\cite{Petsiuk2018rise} and GRAD-CAM~\cite{selvaraju2020grad} on our proposed imageNet eye-tracking data. We can observe that, SIDU outperforms GRAD-CAM and RISE in all the three metrics. Therefore, we can conclude that SIDU explanations are a closer match with the human explanations (heatmaps) for recognizing the object class. This is further illustrated by example image explanation in Figure.~\ref{fig:comparisions of visual explanation with experts} on page~\pageref{fig:comparisions of visual explanation with experts}.
\begin{table} [t!]
\small
\caption{saliency maps of XAI methods with eye fixation maps. }\label{tab:eye_tracking_comparisions}
\begin{centering}
\begin{tabular}{|c|c|c|c|}
\hline 
METHODS & mean KL-DIV$\downarrow$ & mean SCC $\uparrow$ & mean AUC $\uparrow$ \tabularnewline
\hline 
\hline 
RISE~\cite{Petsiuk2018rise}  & 8.4384 & 0.1967  & 0.6385 \tabularnewline
\hline
GRAD-CAM~\cite{selvaraju2020grad} & 9.7892 & 0.2711 & 0.6828   \tabularnewline
\hline 
SIDU  & \textbf{4.3027}  & \textbf{0.3314} & \textbf{0.7708}   \tabularnewline
\hline
\end{tabular}
\par\end{centering}
\end{table}

\subsection{Application-Grounded evaluation} \label{sec:application grouded}

Application-Grounded evaluation involves conducting experiments within a real application to assess the trust of the black-box models. 
We choose an medical case as a test application where we use the task of retinal fundus image quality assessment~\cite{SATYA_VISAPP}. The application is used for screening for retinal diseases, where poor-quality retinal images do not allow an accurate medical diagnosis. Generally, in sensitive domains such as clinical settings, the domain experts ( here clinicians) are skeptical in supporting explanations generated by AI diagnostic tools in cases involving high risk.  

\begin{figure*}[t!]
\begin{centering}
\includegraphics[width= 0.22\textwidth, height = 2.4cm]{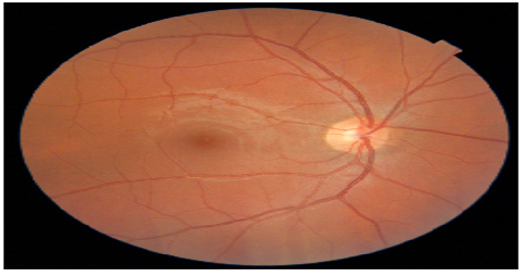}
\hspace{0.01\columnwidth}
\includegraphics[width= 0.22\textwidth, height = 2.4cm]{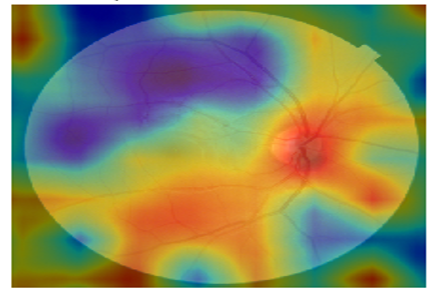}
\hspace{0.01\columnwidth}
\includegraphics[width= 0.22\textwidth, height = 2.4cm]{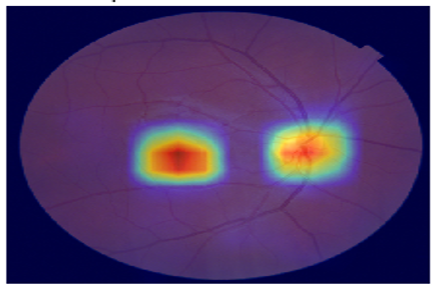}
\hspace{0.01\columnwidth}
\includegraphics[width= 0.22\textwidth, height = 2.4cm]{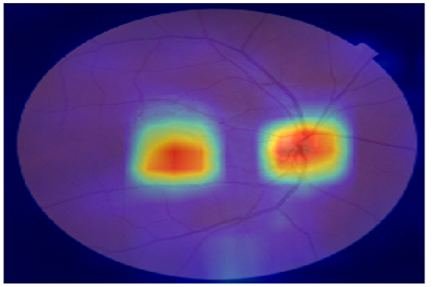}
\par\end{centering}
\begin{centering}
\vspace{1mm}
 
\par\end{centering}
\begin{centering}
\includegraphics[width= 0.22\textwidth, height = 2.4cm]{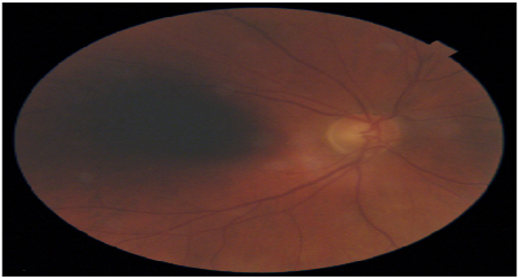}
\hspace{0.01\columnwidth}
\includegraphics[width= 0.22\textwidth, height = 2.4cm]{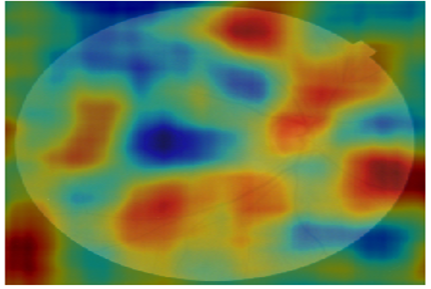}
\hspace{0.01\columnwidth}
\includegraphics[width= 0.22\textwidth, height = 2.4cm]{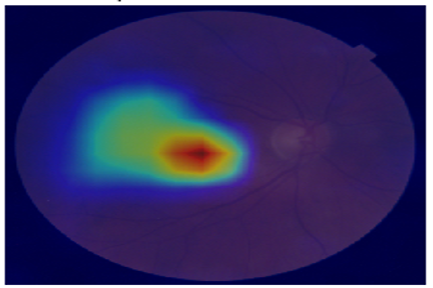}
\hspace{0.01\columnwidth}
\includegraphics[width= 0.22\textwidth, height = 2.4cm]{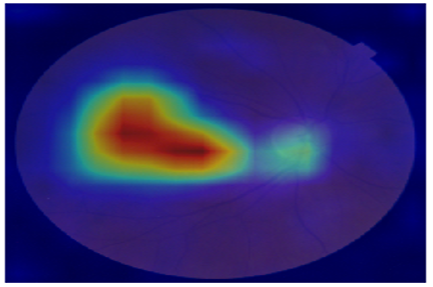}
\par\end{centering}
\centering{}
\scriptsize{(a) Original Image~~~~~~~\hspace{0.50cm}(b) RISE\hspace{1.5cm}(c) GRAD-CAM \hspace{1.5cm} (d)~SIDU}
\caption{ The visual explanation of Good (Top)~/~Bad (Bottom) quality eye fundus images ~~$<$(B), (C), (D)$>$~~ from  RFIQA dataset by RISE, GRAD-CAM and the SIDU method with ResNet50 as the base network. In the real scenario,~the doctors observed the visibility of the optical disc and macular regions in a good quality image ($1^{st}$ image, $1^{st}$ row) corresponding to the region highlighted in the visual explanation heatmap of the proposed method. The bad quality image ($2^{nd}$ image, $2^{nd}$ row) is due to the shadow which is observed near to the center of the image (optical disc),~i.e., exactly the region highlighted by the proposed method.} 
\label{fig:f1s_medical}
\end{figure*}
In our experimental setup at a local hospital, two ophthalmologists participated in testing to evaluate which visual explanation resulted in more trust and further aligns with actual physical examination performed in the clinic. This experiment assesses the effectiveness of the proposed method in terms of localizing the exact region for predicting the retinal fundus image quality with respect to state-of-the-art methods. Here, the generated visual explanation heatmaps in the RISE algorithm were used for comparison. We follow the similar setting as discussed in \cite{selvaraju2020grad}, i.e., using both the proposed SIDU method and the RISE method, visual explanation heatmaps of 100 retinal fundus images for two classes of ‘Good’ and ‘Bad’ quality were recorded. The explanation methods used the trained model as described in section~\ref{sec:functionally_grounded} for explaining the prediction of the retina fundus images. Neither of the ophthalmologists had prior knowledge about any explanation model presented to them. The two explanations methods are labelled as either method I or method II to participants involved in experiments. The participants can opt for “both” methods if they feel that both explanations are rather similar. Therefore, each ophthalmologist will have three different options for every test image. Once the ophthalmologist determined which method better localizes the regions of interest (good/bad quality regions) for each image, we then calculated the relative frequency of each outcome per total retinal fundus image. Table~\ref{t2} on page \pageref{t2} summarizes the results of the two methods evaluated by the experts (with an ophthalmologists). We observed that, in the case of the first ophthalmologist, the RISE explanation map was selected with the relative frequency of $0.02$, the proposed method, SIDU with $0.84$ and $0.14$ being the same. For the second ophthalmologist, the relative frequencies are  $0.05$, $0.93$ and $0.02$, respectively. Therefore, the experiments conclude that, the proposed method gains greater trust from both ophthalmologists and the visual explanations in Figure.~\ref{fig:f1s_medical} further supports this claim.

\begin{table} [t!]
\small
\caption{Expert level evaluation of XAI methods on medical RFIQA dataset . }\label{t2}
\begin{centering}
{\setlength{\extrarowheight}{15pt}}
\begin{tabular}{|c|c|c|}
\hline 
METHODS & Expert I & Expert II  \tabularnewline
\hline 
\hline 
RISE~\cite{Petsiuk2018rise} (Method I) & 0.02 & 0.05 \tabularnewline
\hline 
SIDU (Method II) & \textbf{0.84}  & \textbf{0.93}   \tabularnewline
\hline
BOTH& 0.14 & 0.02   \tabularnewline
\hline
\end{tabular}
\par\end{centering}
\end{table}

\subsection{Effect of Adversarial Noise on XAI methods} \label{sec:adv_noise_experiments}
Despite the success in many applications of AI, recent studies find that Deep Learning is against well designed input samples know as adversarial examples poses a major challenge~\cite{Ann2020}. Adversarial examples are carefully perturbed versions of the original data that successfully fool a classifier. In the image domain, for example, adversarial examples are images that have no visual difference from natural images, but that lead to different classification results. How resilient different XAI algorithms are towards adversarial examples is a largely overlooked topic. In this subsection we therefore investigate exactly that.

\begin{figure*}
\begin{centering}
\includegraphics[width= 0.20\textwidth, height = 2.5cm]{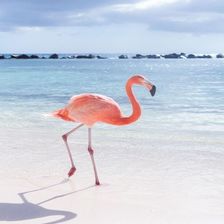} 
\hspace{0.01\columnwidth}
\includegraphics[width= 0.20\textwidth, height = 2.5cm]{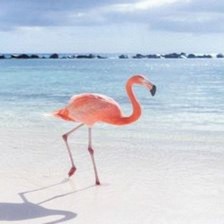}
\hspace{0.01\columnwidth}
\includegraphics[width= 0.20\textwidth, height = 2.5cm]{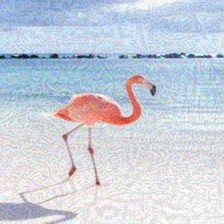}
\hspace{0.01\columnwidth}
\includegraphics[width= 0.20\textwidth, height = 2.5cm]{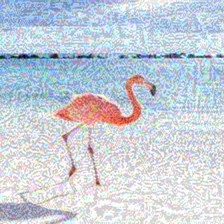}

\par\end{centering}
\begin{centering}
\vspace{1mm}
\par\end{centering}
\centering{}
\scriptsize{\hspace{2mm} (a) Original image\hspace{0.9cm} b)$\epsilon= 0.007$\hspace{0.9cm} ~~~~~~(c)$\epsilon= 0.05$\hspace{0.9cm}~~~~~~(d)  $\epsilon= 0.01$  \hspace{1.5cm}}
\vspace{-.3cm}
\caption{Example of a natural image 'Flamingo' in its original form and also with three different levels of FGSM noise, together with the corresponding predictions 'American egret', 'Nematode' and 'Nematode'.}
\label{fig:noises}
\end{figure*}

To perform this experiment, we choose one the most successful white box attacks, namely, gradient based attacks. Fast Gradient Sign Method (FGSM)~\cite{goodfellow} and Projected Gradient Descent (PGD)~\cite{pgd} are the examples of such attacks. PGD is an iterative application of FGSM such that the process of PGD is more complex and time consuming. Therefore, the Fast Gradient Sign Method (FGSM) was selected because of its simplicity and effectiveness. The adversarial image is generated using FGSM by adding noise to an original image.  The direction of this noise is the same as the gradient of the cost with respect to the input data. 
The amount of noise can be controlled by a coefficient, $\epsilon$. By applying this coefficient properly, it will change the model predictions and it is undetectable to  a human observer. 
Figure~\ref{fig:noises} shows the different levels of FGSM adversarial noise added to an original image. Two different experiments were conducted using adversarial noise to demonstrate the effectiveness of SIDU, compared to the state-of-the-art methods RISE and GRAD-CAM. The experiments are described in the following.

\subsubsection{How do XAI method visual explanations heatmaps of adversarial examples deviate from human eye-fixation heatmaps?}
In this experiment, we analysed how robust the XAI methods are against an adversarial attack in terms of generating reliable explanations. Reliable  visual explanations are defined in terms of resemblance to the human eye-fixation heatmaps. To conduct this experiments we choose the same pre-trained ResNet-50 model used in section~\ref{sec:functionally_grounded}. We first applied the FGSM noise with different epsilon levels to the dataset of 100 images collected from Imagenet validation set as described in section~\ref{sec:eye_tacking_data}. We choose three different optimal noise coefficients between  $0$ and $1$, with the chosen valued being are $\epsilon = 0.007$, $\epsilon = 0.05$ and $\epsilon = 0.1$.  These values were considered optimal because they are sufficient enough to pass unnoticeable by the human eye. We extracted the visual explanations heatmaps using the proposed method SIDU, RISE~\cite{Petsiuk2018rise} and GRAD-CAM~\cite{selvaraju2020grad}. The heatmaps generated by SIDU, RISE and GRAD-CAM methods were finally compared with human generated visual explanations using the eye-tracker as described in section~\ref{sec:eye_tacking_data} using the three evaluation metrics AUC, SCC and KL-DIV. Table~\ref{tab:adv_org_with_adv_noises_merged} on page \pageref{tab:adv_org_with_adv_noises_merged}, summarizes the mean AUC, SCC and KL-DIV results. From the table it can be observed that SIDU outperforms GRAD-CAM and RISE for different levels of adversarial noise with all the three evaluation metrics. We also observe that, the performance of XAI methods decrease with all the three metrics with the increase in adversarial noise to the original images. From this it can concluded that the proposed method (SIDU) has higher robustness to adversarial noise than RISE or GRAD-CAM, as is visually evident in the Figures~\ref{fig:comparisions of visual explanation adversarial noise}. We see that SIDU localizes the entire actual object class after adding the three different levels of adversarial noise, whereas the other methods completely loose the actual object class localization after adding the noise. 


\subsubsection{How do visual explanation maps from adversarial examples deviate
from original visual explanation maps?}

\begin{figure*}[t!]
\begin{centering}

\includegraphics[width= 0.15\textwidth, height = 2.4cm]{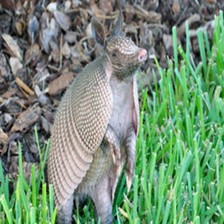}
\hspace{0.01\columnwidth}
\includegraphics[width= 0.15\textwidth, height = 2.4cm]{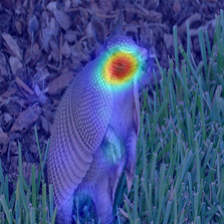}
\hspace{0.01\columnwidth}
\includegraphics[width= 0.15\textwidth, height = 2.4cm]{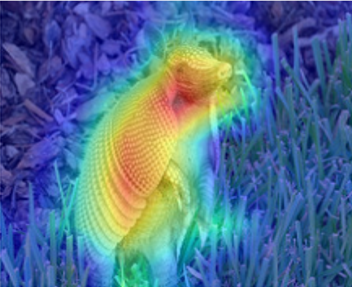}
\hspace{0.01\columnwidth}
\includegraphics[width= 0.15\textwidth, height = 2.4cm]{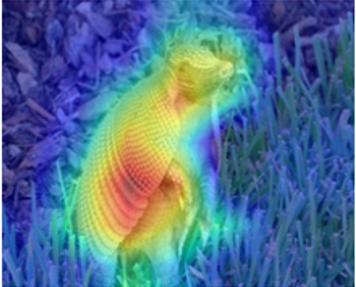}
\hspace{0.01\columnwidth}
\includegraphics[width= 0.15\textwidth, height = 2.4cm]{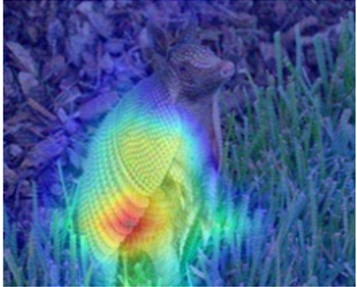}
\hspace{0.01\columnwidth}
\includegraphics[width= 0.15\textwidth, height = 2.4cm]{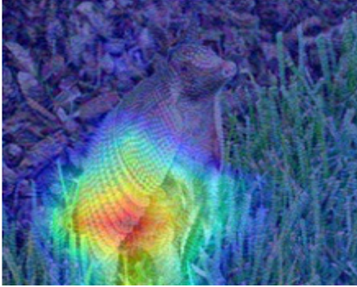}
\par\end{centering}
\begin{centering}
\vspace{1mm}
 
\par\end{centering}
\begin{centering}

\includegraphics[width= 0.15\textwidth, height = 2.4cm]{orginal_adv/adv_0.007_noise21_mongoose.jpg}
\hspace{0.01\columnwidth}
\includegraphics[width= 0.15\textwidth, height = 2.4cm]{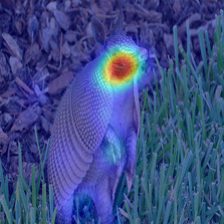}
\hspace{0.01\columnwidth}
\includegraphics[width= 0.15\textwidth, height = 2.4cm]{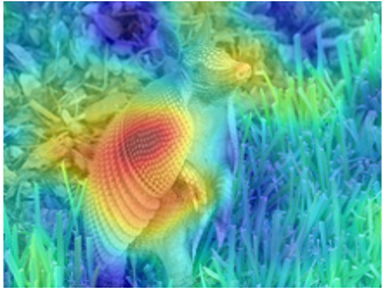}
\hspace{0.01\columnwidth}
\includegraphics[width= 0.15\textwidth, height = 2.4cm]{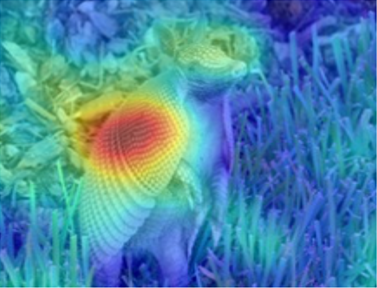}
\hspace{0.01\columnwidth}
\includegraphics[width= 0.15\textwidth, height = 2.4cm]{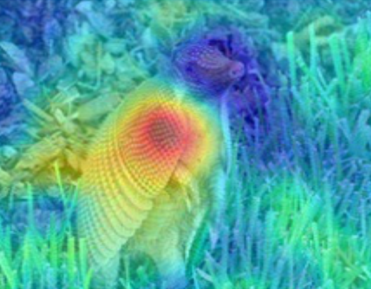}
\hspace{0.01\columnwidth}
\includegraphics[width= 0.15\textwidth, height = 2.4cm]{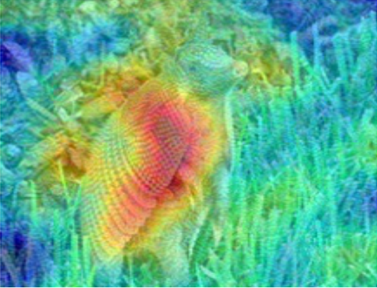}
\par\end{centering}
\begin{centering}
\vspace{1mm}
 
\par\end{centering}
\begin{centering}

\includegraphics[width= 0.15\textwidth, height = 2.4cm]{orginal_adv/adv_0.007_noise21_mongoose.jpg}
\hspace{0.01\columnwidth}
\includegraphics[width= 0.15\textwidth, height = 2.4cm]{adv_org_eye_tracker/image21eye.jpg}
\hspace{0.01\columnwidth}
\includegraphics[width= 0.15\textwidth, height = 2.4cm]{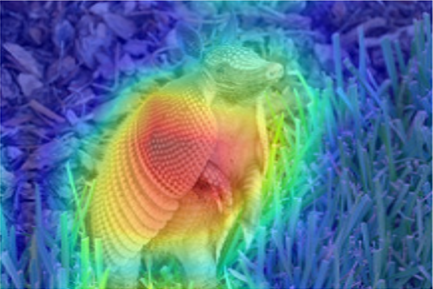}
\hspace{0.01\columnwidth}
\includegraphics[width= 0.15\textwidth, height = 2.4cm]{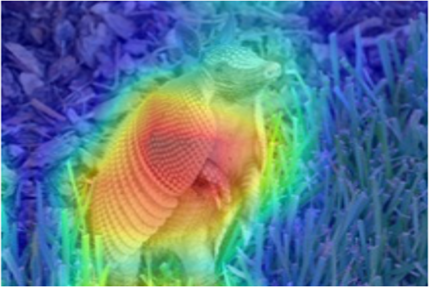}
\hspace{0.01\columnwidth}
\includegraphics[width= 0.15\textwidth, height = 2.4cm]{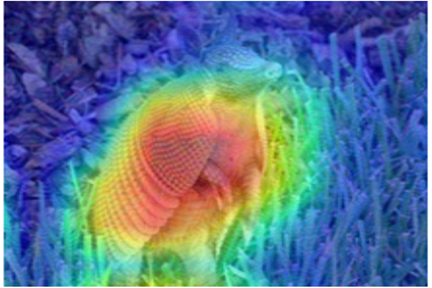}
\hspace{0.01\columnwidth}
\includegraphics[width= 0.15\textwidth, height = 2.4cm]{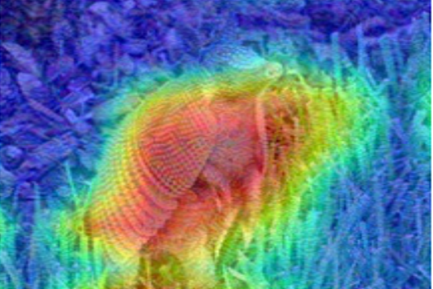}
\par\end{centering}
\begin{centering}
\vspace{1mm}
 
\par\end{centering}

\centering{}
\scriptsize{(a)\hspace{1.6cm}(b)\hspace{1.7cm}(c)\hspace{1.5cm}(d)\hspace{1.9
cm}(e)\hspace{1.5cm}(f)}
\caption{Comparison of XAI visual explanation with different levels of FGSM noise with human visual explanation (heatmaps). The generated heatmaps on adversarial noise levels $\epsilon= 0.007, 0.5,  0.1$. in $3^{rd}$, $4^{th}$ and $5^{th}$  columns by the GRAD-CAM, RISE and SIDU, respectively.  (a) Original Image (b) Eye-tracker (c) $\epsilon = 0 $ (d) $\epsilon = 0.007$ (e) $\epsilon = 0.05$ (f) $\epsilon = 0.1$ } \label{fig:comparisions of visual explanation adversarial noise}
\end{figure*}
In this experiment, we analyse how the visual explanation from adversarial noise added examples of XAI methods deviate from  the original images visual explanation maps.  To conduct this experiments we choose the same pre-trained ResNet-50 model used in section~\ref{sec:functionally_grounded}.  We first applied the FGSM noise with different epsilon levels to the dataset of 100 images collected from Imagenet validation set as described in section~\ref{sec:eye_tacking_data}. We choose one noise level $\epsilon = 0.1$ for these experiments. We extract the visual explanations heatmaps using the proposed method (SIDU), RISE~\cite{Petsiuk2018rise} and GRAD-CAM~\cite{selvaraju2020grad} as applied to the original images without noise and with noise $\epsilon = 0.1$. The heatmaps generated by SIDU, RISE and GRAD-CAM methods are finally compared with the original image visual explanations to see  adversarial noise added images are deviated from the original ones by using the three evaluation metrics AUC, SCC and KL-DIV. Table~\ref{tab:adv_org_with_adv} summarizes the mean AUC, SCC and KL-DIV results obtained by the XAI methods. From the table we can observe that, SIDU outperforms GRAD-CAM and RISE for all the three evaluation metrics. From Figure~\ref{fig:comparisions of visual explanation adversarial noise}, it can be observed that the propose method(SIDU) doesn't deviate in its localizing of the object class that is responsible for the prediction. Therefore, from these two adversarial noise experiments it can be concluded that the proposed method exhibits higher robust against adversarial noise. 
\newcommand*{\MyIndent}{\hspace*{2cm}}
\renewcommand{\multirowsetup}{\centering} 
{\setlength{\extrarowheight}{30pt}}
\begin{table*}[t!]
\caption{Visual explanation heatmaps from adversarial noise $\epsilon$ with eye fixation heatmaps. }\label{tab:adv_org_with_adv_noises_merged}
\Huge
\begin{center}
\begin{adjustbox}{max width=\textwidth}
\def\arraystretch{2}
\begin{tabular}{|
>{\columncolor[HTML]{C0C0C0}}c |c|c|c|c|c|c|c|c|c|}
\hline
\cellcolor[HTML]{C0C0C0}                                                                           & \multicolumn{3}{c|}{$\epsilon= 0.007$}                                                                     & \multicolumn{3}{c|}{$\epsilon= 0.5$}                                                                     & \multicolumn{3}{c|}{$\epsilon= 0.1$}                                                                     \\ \cline{2-10} 
\multirow{-2}{*}{\cellcolor[HTML]{C0C0C0}\begin{tabular}[c]{@{}c@{}}XAI Methods \\ \end{tabular}} & \multicolumn{1}{l|}{mean KL-DIV} & \multicolumn{1}{l|}{mean SCC} & \multicolumn{1}{l|}{mean AUC} & \multicolumn{1}{l|}{mean KL-DIV} & \multicolumn{1}{l|}{mean SCC} & \multicolumn{1}{l|}{mean AUC} & \multicolumn{1}{l|}{mean KL-DIV} & \multicolumn{1}{l|}{mean SCC} & \multicolumn{1}{l|}{mean AUC} \\ \hline
RISE~\cite{Petsiuk2018rise}  & 8.0547  & 0.2121 & 0.6526  & 9.3305 & 0.1995 & 0.6380 & 9.1246 & 0.2068  & 0.6461 \\\hline
GRAD-CAM~\cite{selvaraju2020grad} & 10.3257 & 0.2530 & 0.6719 & 11.6447 & \textbf{0.2229} &  0.6431 & 12.3077 & 0.2112 & 0.6281 \\ \hline
SIDU & \textbf{4.3785} & \textbf{0.3309} & \textbf{0.7689} & \textbf{4.8492}  & \textbf{0.2929} & \textbf{0.7397} & \textbf{4.2239}  & \textbf{0.2817} & \textbf{0.7364}     \\ \hline
\end{tabular}
\end{adjustbox}
\end{center}
\end{table*}

\begin{table} [t!]
\small
\caption{Visual explanation heatmaps from adversarial examples and their deviation
from original visual explanation heatmaps. }\label{tab:adv_org_with_adv}
\begin{centering}
{\setlength{\extrarowheight}{12pt}}
\begin{tabular}{|c|c|c|c|}
\hline 
METHODS & mean KL-DIV$\downarrow$ & mean SCC $\uparrow$ & mean AUC $\uparrow$ \tabularnewline
\hline 
\hline 
RISE~\cite{Petsiuk2018rise}  & 9.6665 & 0.2385 & 0.6133 \tabularnewline
\hline
GRAD-CAM~\cite{selvaraju2020grad} & 10.0077 & 0.4061 & 0.6875   \tabularnewline
\hline 
SIDU  & \textbf{2.4924}  & \textbf{0.6488} & \textbf{0.8347}   \tabularnewline
\hline
\end{tabular}
\par\end{centering}
\end{table}




\section{Conclusion and Future work} \label{sec:conclusion}
In this work, a novel method titled 'Similarity Difference and Uniqueness' method is proposed for explaining the CNN model.Specifically, the investigations were of visual predictions in a form of heatmap through feature activation maps of the last convolution layers in the model. The proposed method is independent of gradients and can effectively localize entire object classes in an image which is responsible for the CNN prediction. The new explanation approach helps in gaining more trust in prediction results of the CNN model by providing further insights to the end-user in sensitive-domains. The effectiveness of our method was validated by conducting three different XAI evaluations methods. These were (1) Application-Grounded (invoking human experts trust in medical domain), (2) Functionally-Grounded (using an automated causal metrics independent of humans) and (3) Human-Grounded evaluation. For the Human-Grounded  evaluation, we proposed a framework for evaluating explainable AI (XAI) methods using an eye-tracker. The framework is designed specifically for evaluating XAI methods using non-experts to understand the human visual perception for recognizing the given object class and compared it with visual explanations of standard well-known CNN models on natural images. Experiments on adversarial examples were also conducted. Results identify our proposed method outperforms compared to state-of-the-art methods. Although comprehensive experimental studies for evaluating XAI methods were conducted, we acknowledge that the experiments involving an eye-tracker are limited only to single-object classification of ten classes. This is due to the fact that there are various methodological challenges associated with eye-tracking (e.g., subject training, hardware calibration, etc) that makes it difficult to access subjects who are willing to participate in data collection for several different scenarios. However, we believe that by demonstrating the great potential of generating valid and reliable explanation via user interaction with an eye-tracker, holds a great value for the research community.
Future work involves extending SIDU to spatio-temporal CNN models to provide visual explanations for video applications tasks such as video classification and action recognition. Further more, exploring the possibility of extending our method to explain decisions made by other neural network architectures (e.g., LSTM), Vision Transformers and in other domains (e.g., Natural Language Processing). We also aim to extend our eye-tracking experimental evaluation on multi-object classification tasks in the future work.~Our code is available at:~~\url{https://github.com/satyamahesh84/SIDU_XAI_CODE}.


\bibliography{mybibfile.bib}

\end{document}